\documentclass[10pt,final,twocolumn,compsoc]{IEEEtran}

\ifCLASSOPTIONcompsoc
  \usepackage[nocompress]{cite}
\else
  \usepackage{cite}
\fi
\ifCLASSINFOpdf
\else
\fi
\usepackage{amsmath}
\usepackage{amsfonts}
\usepackage{amssymb}
\usepackage{amsthm}
\usepackage{eqnarray}
\usepackage{graphicx}
\usepackage{caption}
\usepackage{multirow}

\newtheorem{thm}{Theorem}
\newtheorem{lem}[thm]{Lemma}
\newtheorem{cor}[thm]{Corollary}

\newcommand{\E}{\mathbb{E}}
\newcommand{\bi}[1]{\mbox{\boldmath{$ #1 $}}}

\begin{document}
\title{Estimating Feature-Label Dependence Using Gini Distance Statistics}
%
%
%
%

\author{Silu Zhang,
        Xin Dang,~\IEEEmembership{Member,~IEEE,}
        Dao Nguyen,
        Dawn Wilkins,~Yixin Chen,~\IEEEmembership{Member,~IEEE}
\IEEEcompsocitemizethanks{\IEEEcompsocthanksitem S. Zhang, D. Wilkins, and Y. Chen are with the Department of Computer and Information Science. X. Dang and D. Nguyen are with the Department of Mathematics, University of Mississippi, University, MS 38677, USA. E-mail: \{szhang6,xdang,dxnguyen,dwilkins,yixin\}@olemiss.edu
}
}

%
%

\markboth{IEEE Transactions on Pattern Analysis and Machine Intelligence,~Vol.~, No.~, ~2019}
{Estimating Feature-Label Dependence Using Gini Distance Statistics}
%



\IEEEtitleabstractindextext{%
\begin{abstract}
Identifying statistical dependence between the features and the label is a fundamental problem in supervised learning. This paper presents a framework for estimating dependence between numerical features and a categorical label using \emph{generalized Gini distance}, an energy distance in reproducing kernel Hilbert spaces (RKHS). Two Gini distance based dependence measures are explored: \emph{Gini distance covariance} and \emph{Gini distance correlation}. Unlike Pearson covariance and correlation, which do not characterize independence, the above Gini distance based measures define dependence as well as independence of random variables. The test statistics are simple to calculate and do not require probability density estimation. Uniform convergence bounds and asymptotic bounds are derived for the test statistics. Comparisons with distance covariance statistics are provided. It is shown that Gini distance statistics converge faster than distance covariance statistics in the uniform convergence bounds, hence tighter upper bounds on both Type I and Type II errors. Moreover, the probability of Gini distance covariance statistic under-performing the distance covariance statistic in Type II error decreases to $0$ exponentially with the increase of the sample size. Extensive experimental results are presented to demonstrate the performance of the proposed method.
\end{abstract}

}

\maketitle

\IEEEdisplaynontitleabstractindextext

%
\IEEEpeerreviewmaketitle

\IEEEraisesectionheading{\section{Introduction}\label{sec:introduction}}

%
%
%
%
\IEEEPARstart{B}{uilding} a prediction model from observations of features and responses (or labels) is a well-studied problem in machine learning and statistics. The problem becomes particularly challenging in a high dimensional feature space. A common practice in tackling this challenge is to reduce the number of features under consideration, which is in general achieved via feature combination or feature selection. 

Feature combination refers to combining high dimensional inputs into a smaller set of features via a linear or nonlinear transformation, e.g. principal component analysis (PCA)~\cite{hotelling33}, independent component analysis (ICA)~\cite{comon94}, curvilinear components analysis\cite{demartines97}, multidimensional scaling (MDS)~\cite{torgerson52}, nonnegative matrix factorization (NMF)~\cite{lee99},  Isomap~\cite{tenenbaum00}, locally linear embedding (LLE)~\cite{roweis00}, Laplacian eigenmaps~\cite{belkin02}, stochastic neighbor embedding (SNE)~\cite{hinton02}, etc. Feature selection, also known as variable selection, aims at choosing a subset of features that is ``relevant'' to the response variable~\cite{john94,blum97,kohavi97}. The work proposed in this article belongs to feature selection. Next, we review work most related to ours. For a more comprehensive survey of this subject, the reader is referred to~\cite{guyon03,gui17,liu05,xiang17}.

\subsection{Related Work}
With a common goal of improving the generalization performance of the prediction model and providing a better interpretation of the underlying process, all feature selection methods are built around the concepts of \emph{feature relevance} and \emph{feature redundancy}. 
\subsubsection{Feature Relevance}
The concept of relevance has been studied in many fields outside machine learning and statistics~\cite{hjorland10}. In the context of feature selection, John et al.~\cite{john94} defined feature relevance via a probabilistic interpretation where a feature and the response variable are irrelevant if and only if they are conditionally independent given any subset of features. Following this definition, Nilsson et al.~\cite{nilsson07} investigated distributions under which an optimal classifier can be trained over a minimal number of features. Although the above definition of relevance characterizes the statistical dependence, testing the conditional dependence is in general a challenge for continuous random variables.

Significant amount of efforts have been devoted to finding a good trade-off between theoretical rigor and practical feasibility in defining dependence measures. Pearson Correlation~\cite{pearson1895} based methods are among the most popular approaches. Stoppiglia et al.~\cite{stoppiglia03} proposed a method for linear models using the squared cosine of the angle between a feature and the response variable. Wei and Billings~\cite{wei07} used a squared correlation function to measure dependence between features. Fan and Lv~\cite{fan08} proposed a correlation ranking method for variable selection within the context of linear models. The method was extended in~\cite{fan09} to handle feature selection in generalized linear models and binary classification. Pearson correlation and its variations are in general straightforward to implement. However, it is sensitive only to linear dependence between two variables. Specifically, Pearson correlation can be zero for dependent random variables.

Other correlation measures have been developed to address the above limitation, among which divergence based approaches have been investigated extensively\cite{battiti94}. In~\cite{iannarilli03}, Kullback-Leibler (KL) divergence was used to define the distance between class conditional density functions. A feature subset is chosen to maximize the distance. In~\cite{novovicova96}, features were selected to maximize class separability, which was defined using a variation of the KL divergence. Javed et al.~\cite{javed12} proposed a feature importance function based on KL distance between the joint distribution of a feature and the response variable and their product distribution. In~\cite{wang99}, the dependence between a feature and the target variable is defined as the ratio of the mutual information between them to the entropy of the target variable. Zhai et al.~\cite{zhai16} applied normalized mutual information to filter out noncontributing features. Maji and Pal~\cite{maji10} introduced a mutual information measure for two fuzzy attribute sets. In~\cite{sindhwani04}, Sindhwani et al. presented a method that maximizes the mutual information between the class labels and classifier output. Naghibi et al.~\cite{naghibi15} considered mutual information between features and class labels in a parallel search algorithm.

As mutual information is hard to evaluate, several alternatives have been suggested. Kwak and Choi discussed the limitations of mutual information and proposed two approximation methods~\cite{kwak02a,kwak02b}. Chow and Huang introduced an estimator for mutual information using pruned Parzen window estimator and quadratic mutual information~\cite{chow05}. Peng et al. presented a max-relevance criterion that approximates mutual information between a feature vector and the class label with the sum of the mutual information between individual features and the label~\cite{peng05}. This avoided estimating high-dimensional joint probability. Lefakis and Fleuret proposed a method to maximize bounds on and approximations to the mutual information between the selected features and the class label~\cite{lefakis16}. In~\cite{ding17}, Ding et al. proposed a robust copula dependence (RCD) measure based on the $L_1$ distance of the coupla density from uniform. Theoretical supports were provided to show that RCD is easier to estimate than mutual information.

The above divergence based approaches treat linear and nonlinear dependence under the same framework. There has been researched on tackling nonlinear dependence via linearization. For example, Song et al. used Hilbert-Schmidt Independence Criterion (HISC) to characterize the feature-label dependence. HISC defines cross-covariance in a reproducing kernel Hilbert space (RKHS)~\cite{song12}. Sun et al.~\cite{sun10} transformed a nonlinear problem into a set of locally linear problems, for which feature weights were learned via margin maximization. Armanfard et al.~\cite{armanfard16} proposed a method to associate an ``optimal" feature subset to regions of the sample space. Yao et al.~\cite{yao17} introduced a feature sparsity score that measures the local structure of each feature.

\subsubsection{Feature Redundancy}
Although one may argue that all features dependent on the response variable are informative, redundant features unnecessarily increase the dimensionality of the learning problem, hence may reduce the generalization performance~\cite{yu04}. Eliminating feature redundancy is, therefore, an essential step in feature selection~\cite{peng05}.

Several methods were proposed to reduce redundancy explicitly via a feature dependence measure. Mitra et al.~\cite{mitra02} formulated feature selection as a feature clustering problem. Through a dependence measure, features are partitioned into clusters of high within-group dependence and low between-group dependence. Redundant features are then removed iteratively from each cluster. Wang et al.~\cite{wang08b} adopted a two-step procedure where important (but redundant) features were chosen according to a relevance measure, a forward selection search was then applied to reduce redundancy. In~\cite{benabdeslem14}, feature redundancy is defined via mutual information between two features. A maximum spanning tree is constructed to eliminate redundancy. Tuv et al.~\cite{tuv09} augmented data with artificial features that were independent of the response variable. All features were ranked by an extended random forest method~\cite{breiman01}. Redundant features were then removed based on the ranking of artificial features. Wu et al.~\cite{wu13} applied a Markov blanket to remove redundant features from streaming data.

There are many methods that formulate feature selection as an optimization problem where redundancy reduction is implicitly achieved via optimizing an objective function. For example, Shah et al.~\cite{shah12} integrated redundancy control into learning a conjunction of decision stumps, for which theoretical performance bounds were derived. Chakraborty and Pal~\cite{chakraborty15} introduced a learning algorithm for multi-layer neural networks where the error function included a penalty term associated with feature dependence. Feature redundancy is controlled by minimizing the error function. Yang and Hu~\cite{yang12} proposed a method that selects a subset of feature by minimizing an estimated Bayes error. In~\cite{krishnapuram04}, feature selection was combined with classifier learning under a maximum a posteriori (MAP) framework. A heavy-tailed prior was adopted to promote feature sparsity. He et al.~\cite{he11} developed an algorithm based on Laplacian regularized least squares model. Features were chosen to minimize the ``size" of the covariance matrix and the expected prediction error. In~\cite{constantinopoulos06}, the parameters of a mixture model, the number of components, and saliency of features were optimized simultaneously. Zhao et al.~\cite{zhao13} proposed a method to select features that preserve sample similarity. The method is able to remove redundant features.

Class separation has been widely used as an objective function in redundancy reduction. Bressan and Vitri\`{a}~\cite{bressan03} showed that under the assumption of class-conditional independence, class separability was simplified to a sum of one dimensional KL divergence. Feature selection is performed via ICA. Wang~\cite{wang08a} introduced a class separability criterion in RKHS. Feature selection is achieved by maximizing the criterion. Zhou et al.~\cite{zhou10} showed that maximizing scatter-matrix based class separability may select discriminative yet redundant features. They proposed a redundancy constrained optimization to address the issue. Cheng et al.~\cite{cheng11} proposed to select the most discriminative feature for classification problems. This led to an optimization problem with maximizing the class separations as the objective.

Optimal feature subset selection was investigated under certain optimization formulations. For example, in~\cite{somol04} a special class of monotonic feature selection criterion functions was considered. Branch and Bound algorithms were developed to search for an optimal feature subset. Under general conditions, however, a popular practice is to include a regularization term in the optimization to control the sparsity of the solution. Mao and Tsang~\cite{mao13} proposed a weighted $1$-norm regularizer that favors sparse solutions. Damodaran et al.~\cite{damodaran17} used HISC in a LASSO optimization framework. Wang et al.~\cite{wang14} investigated sparsity regularization for feature selection under an online learning scenario where a small and fixed number of features were maintained. Feature selection for multimodal data was studied in~\cite{wang16}. Multimodal data was projected to a common subspace. The projection and feature selection were obtained simultaneously by solving an optimization problem. Zeng and Cheung~\cite{zeng11} applied a sparse-promoting penalty in learning feature weights in an unsupervised setting. Ren et al.~\cite{ren18} investigated feature selection using a generalized LASSO model. In~\cite{barbu17} the idea of annealing was combined with feature selection using a sparsity favoring loss function. The method gradually removes the most irrelevant features.

\subsection{An Overview of the Proposed Approach}
For problems of large scale (large sample size and/or high feature dimension), feature selection is commonly performed in two steps. A subset of candidate features are first identified via a screening~\cite{fan08} (or a filtering~\cite{guyon03}) process based upon a predefined ``importance" measure that can be calculated efficiently. The final collection of features are then chosen from the candidate set by solving an optimization problem. Usually, this second step is computationally more expensive than the first step. Hence for problems with very high feature dimension, identifying a subset of ``good" candidate features, thus reducing the computational cost of the subsequent optimization algorithm, is essential.

The work presented in this article aims at improving the feature screening process via a new dependence measure. Sz\'{e}kely et al.~\cite{szekely07,szekely09} introduced distance covariance and distance correlation, which extended the classical bivariate product-moment covariance and correlation to random vectors of arbitrary dimension. Distance covariance (and distance correlation) characterizes independence: it is zero if and only if the two random vectors are independent. Moreover, the corresponding statistics are simple to calculate and do not require estimating the distribution function of the random vectors. These properties make distance covariance and distance correlation particularly appealing to the dependence test, which is a crucial component in feature selection~\cite{li12,berrendero16}.

Although distance covariance and distance correlation can be extended to handle categorical variables using a metric space embedding~\cite{lyons13}, Gini distance covariance and Gini distance correlation~\cite{dang18} provide a natural alternative to measuring dependence between a numerical random vector and a categorical random variable. In this article, we investigate selecting informative features for supervised learning problems with numerical features and a categorical response variable using Gini distance covariance and Gini distance correlation. The contributions of this paper are given as follows:
\begin{itemize}
    \item \emph{Generalized Gini distance covariance and Gini distance correlation.} We extend Gini distance covariance and Gini distance correlation to RKHS via positive definite kernels. The choice of kernel not only brings flexibility to the dependence tests, but also makes it easier to derive theoretical performance bounds on the tests.
    \item \emph{Simple dependence tests.} Gini distance statistics are simple to calculate. We prove that when there is dependence between the feature vector and the response variable, the probability of Gini distance covariance statistic under-performing distance covariance statistic approaches $0$ with the growth of the sample size.
    \item \emph{Uniform convergence bounds and asymptotic analysis.} Under the bounded kernel assumption, we derive uniform convergence bounds for both Type I and Type II errors. Compared with distance covariance and distance correlation statistics, the bounds for Gini distance statistics are tighter. Asymptotic analysis is also presented.
\end{itemize}

\subsection{Outline of the Paper}
The remainder of the paper is organized as follows: Section~\ref{ginidistance} motivates Gini distance covariance and Gini distance correlation from energy distance. We then extend them to RKHS and present a connection between generalized Gini distance covariance and generalized distance covariance. Section~\ref{dependence_test} provides estimators of Gini distance covariance and Gini distance correlation. Dependence tests are developed using these estimators. We derive uniform convergence bounds for both Type I and Type II errors of the dependence tests. In Section~\ref{connections} we present connections with dependence tests using distance covariance. Asymptotic results are given in Section~\ref{ab}. We discuss several algorithmic issues in Section~\ref{discussions}. In Section~\ref{experiments}, we explain the extensive experimental studies conducted and demonstrate the results. We conclude and discuss possible future work in Section~\ref{conclusions}.

\section{Gini Distance Covariance and Correlation}\label{ginidistance}
In this section, we first present a brief review of the energy distance. As an instance of the energy distance, Gini distance covariance is introduced to measure dependence between numerical and categorical random variables. Gini distance covariance and correlation are then generalized to reproducing kernel Hilbert spaces (RKHS) to facilitate convergence analysis in Section~\ref{dependence_test}. Connections with distance covariance are also discussed.

\subsection{Energy Distance}
Energy distance was first introduced in~\cite{szekely04,szekely05,baringhaus04} as a measure of statistical distance between two probability distributions with  finite first order moments. The energy distance between the $q$-dimensional independent random variables $X$ and $Y$ is defined as~\cite{szekely13}
\begin{equation}~\label{energyDistance}
    \mathcal{E}(X,Y) = 2 \mathbb{E}|X-Y|_q - \mathbb{E}|X-X'|_q - 
    \mathbb{E}|Y-Y'|_q,
\end{equation}
where $|\cdot|_q$ is the Euclidean norm in $\mathbb{R}^q$, $\mathbb{E}|X|_q + \mathbb{E}|Y|_q < \infty$, $X'$ is an iid copy of $X$, and $Y'$ is an iid copy of $Y$. 

Energy distance has many interesting properties. It is scale equivariant: for any $a \in \mathbb{R}$,
$$
\mathcal{E}(aX,aY) = |a|\mathcal{E}(X,Y).
$$
It is rotation invariant: for any rotation matrix $\mathbf{R} \in \mathbb{R}^{q\times q}$
$$
\mathcal{E}(\mathbf{R}X,\mathbf{R}Y) = \mathcal{E}(X,Y).
$$
Test statistics of an energy distance are in general relatively simple to calculate and do not require density estimation (Section~\ref{dependence_test}). Most importantly, as shown in~\cite{szekely05}, if $\varphi_X$ and $\varphi_Y$ are the characteristic functions of $X$ and $Y$, respectively, the energy distance (\ref{energyDistance}) can be equivalently written as
\begin{equation}~\label{energyDistanceL2}
   \mathcal{E}(X,Y) = c(q)\int_{\mathbb{R}^q} \frac{\left[ \varphi_X(x) - \varphi_Y(x)\right]^2}{|x|_q^{q+1}} dx,
\end{equation}
where $c(q)>0$ is a constant only depending on $q$. Thus $\mathcal{E} \ge 0$ with equality to zero if and only if $X$ and $Y$ are identically distributed. The above properties make energy distance especially appealing to testing identical distributions (or dependence).

\subsection{Gini Distance Covariance and Gini Distance Correlation}\label{ginidist}
Gini distance covariance was proposed in~\cite{dang18} to measure dependence between a numerical random variable $X \in \mathbb{R}^q$ from function $F$ (cumulative distribution function, CDF) and a categorical variable $Y$ with $K$ values $L_1,...,L_K$. If we assume the categorical distribution $P_Y$ of $Y$ is $\mathrm{Pr}(Y = L_k) = p_k$ and the conditional distribution of $X$ given $Y=L_k$ is $F_k$, the marginal distribution of $X$ is 
$$
F(x) = \sum_{k=1}^K p_k F_k(x).
$$
When the conditional distribution of $X$ given $Y$ is the same as the marginal distribution of $X$, $X$ and $Y$ are independent, i.e., there is no correlation between them. However, when they are dependent, i.e., $F \neq F_k$ for some $k$, the dependence can be measured through the difference between the marginal distribution $F$ and conditional distribution $F_k$. 

This difference is measured by Gini distance covariance, $\mathrm{gCov}(X,Y)$, which is defined as the expected weighted $L_2$ distance between characteristic functions of the conditional and marginal distribution (if the expectation is finite):
\begin{equation*}\label{gDistCovI}
\mathrm{gCov}(X,Y): = c(q)\sum_{k=1}^K p_k \int_{\mathbb{R}^q} \frac{[\varphi_k(x) - \varphi(x)]^2}{|x|_q^{q+1}} d x,
\end{equation*}
where $c(q)$ is the same constant as in (\ref{energyDistanceL2}), $\varphi_k$ and $\varphi$ are the characteristic functions for the conditional distribution $F_k$ and marginal distribution $F$, respectively. It follows immediately that $\mathrm{gCov}(X,Y)=0$ mutually implies independence between $X$ and $Y$. Based on (\ref{energyDistance}) and (\ref{energyDistanceL2}), the Gini distance covariance is clearly a weighted energy distance, hence can be equivalently defined as
\begin{align}
&\mathrm{gCov}(X,Y) \nonumber \\
&= \sum_{k=1}^{K} p_k \left[ 2 \mathbb{E}|X_k-X|_q - 
    \mathbb{E}|X_k-{X_k}'|_q - \mathbb{E}|X-X'|_q\right], \label{gDistCov}
\end{align}
where $(X_k,{X_k}')$ and $(X,X')$ are independent pair variables from $F_k$ and $F$, respectively. 

Gini distance covariance can be standardized to have a range of $[0,1]$, a desired property for a correlation measure. The resulting measure is called Gini distance correlation, denoted by $\mathrm{gCor}(X,Y)$, which is defined as
\begin{align}
    &\mathrm{gCor}(X,Y) \nonumber \\
    &=\frac{\sum_{k=1}^{K} p_k \left[ 2 \mathbb{E}|X_k-X|_q -\mathbb{E}|X_k-{X_k}'|_q - \mathbb{E}|X-X'|_q\right]}{\mathbb{E}|X-X'|_q}.\label{gDistCor}
\end{align}
provided that $\mathbb{E}|X|_q + \mathbb{E}|X_k|_q < \infty$ and $F$ is not a degenerate distribution. Gini distance correlation satisfies the following properties~\cite{dang18} . 
\begin{enumerate}
\item $0\leq \mathrm{gCor}(X,Y) \leq 1$.
\item $\mathrm{gCor}(X,Y) =0$ if and only if $X$ and $Y$ are independent.
\item $\mathrm{gCor}(X,Y)=1 $ if and only if $F_k$ is a single point mass distribution almost surely for all $k = 1,...,K$.
\item $\mathrm{gCor}(a\mathbf{R}X+b,Y) = \mathrm{gCor}(X,Y)$ for all $a \ne 0$, $b\in \mathbb{R}^q$, and any orthonormal matrix $\mathbf{R} \in \mathbb{R}^{q \times q}$.
\end{enumerate}
Property $2$ are especially useful in testing dependence.

\subsection{Gini Distance Statistics in RKHS}~\label{gDistRKHS}
Energy distance based statistics naturally generalizes from a Euclidean space to metric spaces~\cite{lyons13}. By using a positive definite kernel (Mercer kernel)~\cite{mercer1909}, distributions are mapped into a RKHS~\cite{smola07} with a kernel induced distance. Hence one can extend energy distances to a much richer family of statistics defined in RKHS~\cite{sejdinovic13}. Let $\kappa: \mathbb{R}^q \times \mathbb{R}^q \rightarrow{\mathbb{R}}$ be a Mercer kernel~\cite{mercer1909}. There is an associated RKHS $\mathcal{H}_{\kappa}$ of real functions on $\mathbb{R}^q$ with reproducing kernel $\kappa$, where the function $d: \mathbb{R}^q \times \mathbb{R}^q \rightarrow{\mathbb{R}}$ defines a distance in $\mathcal{H}_\kappa$,
\begin{equation}\label{kdist}
    d_\kappa(x,x') = \sqrt{\kappa(x,x) + \kappa(x',x') - 2 \kappa(x,x')}.
\end{equation}
Hence Gini distance covariance and Gini distance correlation are generalized to RKHS, $\mathcal{H}_\kappa$, as
\begin{align}
    &\mathrm{gCov}_\kappa(X,Y) \nonumber \\
    &= \sum_{k=1}^{K} p_k \left[ 2 \mathbb{E}d_\kappa(X_k,X) - 
    \mathbb{E}d_\kappa(X_k,{X_k}') - \mathbb{E}d_\kappa(X,X')\right],\label{gCovk}\\
    &\mathrm{gCor}_\kappa(X,Y) \nonumber \\
    & = \frac{\sum_{k=1}^{K} p_k \left[ 2 \mathbb{E}d_\kappa(X_k,X) - 
    \mathbb{E}d_\kappa(X_k,{X_k}') - \mathbb{E}d_\kappa(X,X')\right]}{\mathbb{E}d_\kappa(X,X')}.\label{gCork}
\end{align}
The choice of kernels allows one to design various tests. In this paper, we focus on bounded translation and rotation invariant kernels. Our choice is based on the following considerations: 
\begin{enumerate}
    \item The boundedness of a positive definite kernel implies the boundedness of the distance in RKHS, which makes it easier to derive strong (exponential) convergence inequalities based on bounded deviations (discussed in Section~\ref{dependence_test});
    \item Translation and rotation invariance is an important property to have for testing of dependence.
\end{enumerate}

Same as in $\mathbb{R}^q$, Gini distance covariance and Gini distance correlation in RKHS also characterize independence, i.e., $\mathrm{gCov}_\kappa(X,Y) = 0$ and $\mathrm{gCor}_\kappa(X,Y)=0$ if and only if $X$ and $Y$ are independent. This is derived as the following from the connection between Gini distance covariance and distance covariance in RKHS. Distance covariance was introduced in~\cite{szekely07} as a dependence measure between random variables $X \in \mathbb{R}^p$ and $Y \in \mathbb{R}^q$. If $X$ and $Y$ are embedded into RKHS's induced by $\kappa_X$ and $\kappa_Y$, respectively, the generalized distance covariance of $X$ and $Y$ is~\cite{sejdinovic13}:
\begin{align}
    &\mathrm{dCov}_{\kappa_X,\kappa_Y}(X,Y) \nonumber\\
    & = \mathbb{E}d_{\kappa_X}(X,X')d_{\kappa_Y}(Y,Y') + \mathbb{E}d_{\kappa_X}(X,X')\mathbb{E}d_{\kappa_Y}(Y,Y') \nonumber\\
    & - 2\mathbb{E}\left[\mathbb{E}_{X'}d_{\kappa_X}(X,X') \mathbb{E}_{Y'}d_{\kappa_Y}(Y,Y')\right].\label{dCovkk}
\end{align}

In the case of $Y$ being categorical, one may embed it using a set difference kernel $\kappa_Y$,
\begin{equation}\label{setdiff}
    \kappa_Y(y,y') = \left\{ \begin{array}{cc}
    \frac{1}{2} &  if \;y = y',\\
    0 & otherwise.
    \end{array} \right.
\end{equation}
This is equivalent to embedding $Y$ as a simplex with edges of unit length~\cite{lyons13}, i.e., $L_k$ is represented by a $K$ dimensional vector of all zeros except its $k$-th dimension, which has the value $\frac{\sqrt{2}}{2}$. The distance induced by $\kappa_Y$ is called the set distance, i.e., $d_{\kappa_Y}(y,y')=0$ if $y=y'$ and $1$ otherwise. Using the set distance, we have the following results on the generalized distance covariance between a numerical and a categorical random variable.
\begin{lem}\label{lem_dcovk}
Suppose that $X \in \mathbb{R}^q$ is from distribution $F$ and $Y$ is a categorical variable with $K$ values $L_1,...,L_K$. The categorical distribution $P_Y$ of $Y$ is $P(Y = L_k) = p_k$ and the conditional distribution of $X$ given $Y=L_k$ is $F_k$, the marginal distribution of $X$ is $F(x) = \sum_{k=1}^K p_k F_k(x).$ Let $\kappa_X: \mathbb{R}^q \times \mathbb{R}^q \rightarrow{\mathbb{R}}$ be a Mercer kernel and $\kappa_Y$ a set difference kernel. The generalized distance covariance $\mathrm{dCov}_{\kappa_X,\kappa_Y}(X,Y)$ is equivalent to
\begin{align}
    &\mathrm{dCov}_{\kappa_X,\kappa_Y}(X,Y) := \mathrm{dCov}_{\kappa_X}(X,Y) \nonumber\\
    & = \sum_{k=1}^{K} p_k^2 \left[2 \mathbb{E}d_{\kappa_X}(X_k,X) - \mathbb{E}d_{\kappa_X}(X_k,{X_k}') - \mathbb{E}d_{\kappa_X}(X,X') \right].\label{dCovk}
\end{align}
\end{lem}
From (\ref{gCovk}) and (\ref{dCovk}), it is clear that the generalized Gini covariance is always larger than or equal to the generalized distance covariance under the set difference kernel and the same $\kappa_X$, i.e.,~\footnote{The inequality holds for Gini covariance and distance covariance as well, i.e., $\mathrm{gCov}(X,Y) \ge \mathrm{dCov}(X,Y)$ where $X \in \mathbb{R}^d$ and $Y$ is categorical. The notations of $\mathrm{gCov}_{\kappa_X}(X,Y)$ and $\mathrm{gCov}_{\kappa}(X,Y)$ are used interchangeably with both $\kappa_X$ and $\kappa$ representing a Mercer kernel.}
\begin{equation}\label{gCov>=dCov}
    \mathrm{gCov}_{\kappa_X}(X,Y) \ge \mathrm{dCov}_{\kappa_X}(X,Y)
\end{equation}
where they are equal if and only if both are $0$, i.e., $X$ and $Y$ are independent. This yields the following theorem. The proof of Lemma~\ref{lem_dcovk} is given in Appendix~\ref{app_a}.
\begin{thm}\label{independence}
For any bounded Mercer kernel $\kappa: \mathbb{R}^q \times \mathbb{R}^q \rightarrow{\mathbb{R}}$, $\mathrm{gCov}_\kappa(X,Y) = 0$ if and only if $X$ and $Y$ are independent. The same result holds for  $\mathrm{gCor}_\kappa(X,Y)$ assuming that the marginal distribution of $X$ is not degenerate.
\end{thm}
\begin{proof}
The proof of the sufficient part for $\mathrm{gCov}_\kappa(X,Y)$ is immediate from the definition (\ref{gCovk}). The inequality (\ref{gCov>=dCov}) suggests that $\mathrm{dCov}_\kappa = 0$ when $\mathrm{gCov}_\kappa = 0$. Hence the proof of the necessary part is complete if we show that $\mathrm{dCov}_\kappa = 0$ implies independence. This is proven as the following.

Let $\mathcal{X}$ and $\mathcal{Y}$ be the RKHS induced by $\kappa$ and the set difference kernel (\ref{setdiff}), respectively, with the associated distance metrics defined according to (\ref{kdist}). $\mathcal{X}$ and $\mathcal{Y}$ are both separable Hilbert spaces~\cite{aronszain50,canu03} as they each have a countable set of orthonormal basis~\cite{mercer1909}. Hence $\mathcal{X}$ and $\mathcal{Y}$ are of strong negative type (Theorem 3.16 in~\cite{lyons13}). Because the metrics on $\mathcal{X}$ and $\mathcal{Y}$ are bounded, the marginals of $(X,Y)$ on $\mathcal{X} \times \mathcal{Y}$ have finite first moment in the sense defined in~\cite{lyons13}. Therefore, $\mathrm{dCov}_\kappa(X,Y) = 0$ implies that $X$ and $Y$ are independent (Theorem 3.11~\cite{lyons13}).

Finally, the proof for $\mathrm{gCor}_\kappa(X,Y)$ follows from the above and the condition that $X$ is not degenerate.
\end{proof}

In the remainder of the paper, unless noted otherwise, we use the default distance function~\footnote{Since any bounded translation and rotation invariant kernels can be normalized to define a distance function with the maximum value no greater than $1$, the results in Section~\ref{gDistRKHS} and Section~\ref{dependence_test} hold for these kernels as well.}
$$
d_\kappa(x,x') = \sqrt{1-e^{-\frac{|x-x'|_q^2}{\sigma^2}}},
$$
induced by a weighted Gaussian kernel, $\kappa(x,x') = \frac{1}{2}e^{-\frac{|x-x'|_q^2}{\sigma^2}}.$ It is immediate that the above distance function is translation and rotation invariant and is
bounded with the range $[0,1)$. Moreover, using Taylor expansion, it is not difficult to show that $\mathrm{gCor}_\kappa$ approaches $\mathrm{gCor}$ when the kernel parameter $\sigma$ approaches $\infty$.

\section{Dependence Tests}\label{dependence_test}
We first present an unbiased estimator of the generalized Gini distance covariance. Probabilistic bounds for large deviations of the empirical generalized Gini distance covariance are then derived. These bounds lead directly to two dependence tests. We also provide discussions on connections with the dependence test using generalized distance covariance. Finally, asymptotic analysis of the test statistics is presented. 

\subsection{Estimation}\label{unbiased}
In Section~\ref{ginidist}, Gini distance covariance and Gini distance correlation were introduced from an energy distance point of view. An alternative interpretation based on Gini mean difference was given in~\cite{dang18}. This definition yields simple point estimators.

Let $X \in \mathbb{R}^q$ be a random variable from distribution $F$. Let $Y \in \mathbb{Y} = \{ L_1,\cdots,L_K\}$ be a categorical random variable with $K$ values and $\mathrm{Pr}(Y = L_k) = p_k \in (0,1)$. The conditional distribution of $X$ given $Y=L_k$ is $F_k$. Let $(X,X')$ and $(X_k,{X_k}')$ be independent pair variables from $F$ and $F_k$, respectively. The Gini distance covariance (\ref{gDistCov}) and Gini distance correlation (\ref{gDistCor}) can be equivalently written as
\begin{eqnarray}
    \mathrm{gCov}(X,Y) & = & \Delta - \sum_{k=1}^{K}{p_k \Delta_k}, \label{gini_gcov_alt}\\
    \mathrm{gCor}(X,Y) & = & \frac{\Delta - \sum_{k=1}^{K}{p_k \Delta_k}}{\Delta}, \label{gini_gcor_alt}
\end{eqnarray}
where $\Delta = \mathbb{E}{|X-X'|_q}$ and $\Delta_k = \mathbb{E}{|X_k - {X_k}'|_q}$ are the Gini mean difference (GMD) of $F$ and $F_k$ in $\mathbb{R}^q$~\cite{gini12,gini14,Koshevoy97}, respectively. This suggests that Gini distance covariance is a measure of between-group variation and Gini distance correlation is the ratio of between-group variation and the total Gini variation. Replacing $|\cdot|_q$ with $d_\kappa(\cdot,\cdot)$ in (\ref{gini_gcov_alt}) and (\ref{gini_gcor_alt}) yields the GMD version of (\ref{gCovk}) and (\ref{gCork}).

Given an iid sample data $\mathcal{D}=\left\{(x_i,y_i) \in \mathbb{R}^q \times \mathbb{Y}: i=1,\cdots,n \right\}$, let $\mathcal{I}_k$ be the index set of sample points with $y_i = L_k$. The probability $p_k$ is estimated by the sample proportion of category $k$, i.e., $\hat{p}_k = \frac{n_k}{n}$ where $n_k = |\mathcal{I}_k| > 2$. The point estimators of the generalized Gini distance covariance and Gini distance correlation for a given kernel $\kappa$ are
\begin{eqnarray}
    \mathrm{gCov}_\kappa^n & := & \hat{\Delta} - \sum_{k=1}^{K} \hat{p}_k \hat{\Delta}_k, \label{gCovEst}\\
    \mathrm{gCor}_\kappa^n & := & \frac{\hat{\Delta}-\sum_{k=1}^{K} \hat{p}_k\hat{\Delta}_k}{\hat{\Delta}}, \label{gCorEst}
\end{eqnarray}
where
\begin{eqnarray}
    \hat{\Delta}_k & = & \binom{n_k}{2}^{-1} \sum_{i<j \in \mathcal{I}_k}{d_\kappa(x_i,x_j)}, \label{deltak}\\
    \hat{\Delta} & = & \binom{n}{2}^{-1} \sum_{i<j}{d_\kappa(x_i,x_j)}. \label{delta}
\end{eqnarray}

\begin{thm}
The point estimator (\ref{gCovEst}) of the generalized Gini distance covariance is unbiased.
\end{thm}
\begin{proof}
Clearly, $\hat{\Delta}_k$ and $\hat{\Delta}$ are unbiased because they are U-statistics of size 2. Also $\hat{p}_k \hat \Delta_k$ is unbiased since  $\E [\hat{p}_k \hat \Delta_k] = \E \E[\hat{p}_k\hat \Delta_k|n_k] = \E [\frac{n_k}{n} \Delta_k] = p_k \Delta_k$. This leads to the unbiasedness of $\mathrm{gCov}_\kappa^n$. 
\end{proof}

\subsection{Uniform Convergence Bounds}\label{ucb}
We derive two probabilistic inequalities, from which dependence tests using point estimators (\ref{gCovEst}) and (\ref{gCorEst}) are established.
\begin{thm}\label{ucb_gCovk}
Let $\mathcal{D}=\left\{(x_i,y_i)\in \mathbb{R}^q \times \mathbb{Y}: i=1,\cdots,n \right\}$ be an iid sample of $(X,Y)$ and $\kappa$ a Mercer kernel over $\mathbb{R}^q \times \mathbb{R}^q$ that induces a distance function $d_\kappa(\cdot,\cdot)$ with bounded range $[0,1)$. For every $\epsilon > 0$,
\begin{eqnarray*}
    \mathrm{Pr}\left[\mathrm{gCov}_\kappa^n - \mathrm{gCov}_\kappa(X,Y) \ge \epsilon \right] & \le & \exp{\left(\frac{-n\epsilon^2}{12.5}\right)},\; \mathrm{and} \\
    \mathrm{Pr}\left[\mathrm{gCov}_\kappa(X,Y) - \mathrm{gCov}_\kappa^n \ge \epsilon \right] & \le & \exp{\left(\frac{-n\epsilon^2}{12.5}\right)}.
\end{eqnarray*}
\end{thm}
\begin{thm}\label{ucb_delta}
Under the condition of Theorem~\ref{ucb_gCovk}, for every $\epsilon > 0$
\begin{eqnarray*}
    \mathrm{Pr}[\hat{\Delta} - \Delta \ge \epsilon] & \le & \exp{\left(\frac{-n\epsilon^2}{2}\right)},\;\mathrm{and}\\
    \mathrm{Pr}[\Delta - \hat{\Delta} \ge \epsilon] & \le & \exp{\left(\frac{-n\epsilon^2}{2}\right)}.
\end{eqnarray*}
\end{thm}

Proofs of Theorem~\ref{ucb_gCovk} and Theorem~\ref{ucb_delta} are given in Appendix~\ref{app_b}. Next we consider a dependence test based on $\mathrm{gCov}_\kappa^n$. Theorem~\ref{independence} shows that $\mathrm{gCov}_\kappa(X,Y) = 0$ mutually implies that $X$ and $Y$ are independent. This suggests the following null and alternative hypotheses:
\begin{eqnarray*}
    H_0 & : & \mathrm{gCov}_\kappa(X,Y) = 0, \\
    H_1 & : & \mathrm{gCov}_\kappa(X,Y) \ge 2cn^{-t},\;c > 0\; \mathrm{and}\; t > 0.
\end{eqnarray*}
The null hypothesis is rejected when $\mathrm{gCov}_\kappa^n \ge c n^{-t}$ where $c > 0$ and $t \in \left(0,\frac{1}{2}\right)$. Next we establish upper bounds for the Type I and Type II errors of the above dependence test.
\begin{cor}\label{upperbound_gini}
Under the conditions of Theorem~\ref{ucb_gCovk}, the following inequalities hold for any $c > 0$ and $t \in \left(0,\frac{1}{2} \right)$:
\begin{align}
    \mathrm{Type\;I} & :  \mathrm{Pr}\left[\mathrm{gCov}_\kappa^n \ge c n^{-t} |H_0\right] \le \exp{\left( -\frac{c^2 n^{1-2t}}{12.5}\right)}, \label{gCovT1} \\
    \mathrm{Type\;II} & :  \mathrm{Pr}\left[\mathrm{gCov}_\kappa^n \le c n^{-t} |H_1\right] \le \exp{\left( -\frac{c^2 n^{1-2t}}{12.5}\right)}. \label{gCovT2}
\end{align}
\end{cor}
\begin{proof}
Let $\epsilon = cn^{-t}$. The Type I bound is immediate from Theorem~\ref{ucb_gCovk}. The Type II bound is derived from the following inequality and Theorem~\ref{ucb_gCovk}.
\begin{align*}
    &\mathrm{Pr}\left[\mathrm{gCov}_\kappa^n \le c n^{-t}|H_1\right] \\
    & \le \mathrm{Pr}\left[c n^{-t} - \mathrm{gCov}_\kappa^n + \mathrm{gCov}_\kappa(X,Y) - 2c n^{-t} \ge 0 | H_1 \right] \\
    & = \mathrm{Pr}\left[\mathrm{gCov}_\kappa(X,Y) - \mathrm{gCov}_\kappa^n \ge c n^{-t} | H_1 \right].
\end{align*}

\end{proof}

A dependence test can also be performed using the empirical Gini distance correlation under the above null and alternative hypotheses with $\mathrm{gCor}_\kappa$ replacing $\mathrm{gCov}_\kappa$. The null hypothesis is rejected when $\mathrm{gCor}_\kappa^n \ge c n^{-t}$ where $c > 0$ and $t \in \left(0, \frac{1}{4}\right)$. Type I and Type II bounds are presented as below.
\begin{cor}
Under the conditions of Theorem~\ref{ucb_gCovk} and Theorem~\ref{ucb_delta} where additionally $\Delta \ge 2 n^{-t}$, the following inequalities hold for any $c > 0$ and $t \in \left(0, \frac{1}{4} \right)$:
\begin{align}
    \mathrm{Type\;I} & : \mathrm{Pr}\left[\mathrm{gCor}_\kappa^n \ge c n^{-t} | H_0\right] \nonumber\\
    &\quad\le \exp{\left(-\frac{c^2 n^{1-4t}}{12.5}\right)} + \exp{\left(-\frac{n^{1-2t}}{2}\right)}, \label{gCorT1} \\
    \mathrm{Type\;II} & : \mathrm{Pr}\left[\mathrm{gCor}_\kappa^n \le c n^{-t} | H_1 \right] \le \exp{\left( -\frac{c^2 n^{1-2t}}{12.5}\right)}. \label{gCorT2}
\end{align}
\end{cor}
\begin{proof}
From (\ref{gCorEst}), we have
\begin{align*}
& \mathrm{Pr}\left[\mathrm{gCor}_\kappa^n \ge c n^{-t} | H_0\right] \\
&\le \mathrm{Pr}\left[ \mathrm{gCov}_\kappa^n \ge c n^{-2t}\;\; \mathrm{OR}\;\; \hat{\Delta} \le n^{-t} | H_0 \right]\\
&\le \mathrm{Pr}\left[ \mathrm{gCov}_\kappa^n \ge c n^{-2t} | H_0 \right] + \mathrm{Pr}\left[ \hat{\Delta} \le n^{-t} | H_0 \right]\\
&\le \mathrm{Pr}\left[ \mathrm{gCov}_\kappa^n \ge c n^{-2t} | H_0 \right] + \mathrm{Pr}\left[ \Delta - \hat{\Delta} \ge n^{-t} | H_0 \right].
\end{align*}
Let $\epsilon_1=cn^{-2t}$ and $\epsilon_2=n^{-t}$. The Type I bound is derived from Theorem~\ref{ucb_gCovk} and Theorem~\ref{ucb_delta}. The boundedness of $d_\kappa(\cdot,\cdot)$ implies that $\hat{\Delta} < 1$. Therefore,
\begin{align*}
& \mathrm{Pr}\left[\mathrm{gCor}_\kappa^n \le c n^{-t} |H_1\right] \\
& \le \mathrm{Pr}\left[ \mathrm{gCov}_\kappa^n \le c n^{-t} | H_1 \right] \\
& \le \mathrm{Pr}\left[\mathrm{gCov}_\kappa(X,Y) - \mathrm{gCov}_\kappa^n \ge c n^{-t} | H_2 \right].
\end{align*}
Hence the Type II bound is given by Theorem~\ref{ucb_gCovk} with
$\epsilon = c n^{-t}$.
\end{proof}

\subsection{Connections to Generalized Distance Covariance} \label{connections}
In Section~\ref{gDistRKHS}, generalized Gini distance covariance is related to generalized distance covariance through (\ref{gCov>=dCov}). Under the conditions of Lemma~\ref{lem_dcovk}, $\mathrm{dCov}_{\kappa_X,\kappa_Y}(X,Y)=0$ if and only if $X$ and $Y$ are independent. Hence dependence tests similar to those in Section~\ref{ucb} can be developed using empirical estimates of
$\mathrm{dCov}_{\kappa_X,\kappa_Y}(X,Y)$. Next, we establish a result similar to Theorem~\ref{ucb_gCovk} for generalized distance covariance. We demonstrate that generalized Gini distance covariance has a tighter probabilistic bound for large deviations than its generalized distance covariance counterpart.

Using the unbiased estimator for distance covariance developed in~\cite{szekely14}, we generalize it to an unbiased estimator for $\mathrm{dCov}_{\kappa_X,\kappa_Y}(X,Y)$ defined in (\ref{dCovkk}). Let $\mathcal{D}=\left\{(x_i,y_i)\in \mathbb{R}^q \times \mathbb{R}^p: i=1,\cdots,n \right\}$ be an iid sample from the joint distribution of $X$ and $Y$. Let $A=(a_{ij})$ be a symmetric, $n \times n$, centered kernel distance matrix of sample $x_1,\cdots,x_n$. The $(i,j)$-th entry of $A$ is
\begin{equation*}
    A_{ij} = 
    \begin{cases}
    a_{ij}-\frac{1}{n-2}a_{i\cdot}-\frac{1}{n-2}a_{\cdot j} + \frac{1}{(n-1)(n-2)}a_{\cdot \cdot}, & i\ne j;\\
    0, & i=j,
    \end{cases}
\end{equation*}
where $a_{ij} = d_{\kappa_X}(x_i,x_j)$, $a_{i\cdot} = \sum_{j=1}^n a_{ij}$, $a_{\cdot j} = \sum_{i=1}^n a_{ij}$, and $a_{\cdot \cdot}=\sum_{i,j=1}^n a_{ij}$. Similarly, using $d_{\kappa_Y}(y_i,y_j)$, a symmetric, $n \times n$, centered kernel distance matrix is calculated for samples $y_1,\cdots, y_n$ and denoted by $B = (b_{ij})$. An unbiased estimator of $\mathrm{dCov}_{\kappa_X,\kappa_Y}(X,Y)$ is given as
\begin{equation}\label{udCov}
    \mathrm{dCov}_{\kappa_X, \kappa_Y}^n = \frac{1}{n(n-3)}\sum_{i\ne j}A_{ij}B_{ij}.
\end{equation}
We have the following result on the concentration of $\mathrm{dCov}_{\kappa_X, \kappa_Y}^n$ around $\mathrm{dCov}_{\kappa_X,\kappa_Y}(X,Y)$. 
\begin{thm}\label{ucb_dCovk}
Let $\mathcal{D}=\left\{(x_i,y_i)\in \mathbb{R}^q \times \mathbb{R}^p: i=1,\cdots,n \right\}$ be an iid sample of $(X,Y)$. Let $\kappa_X: \mathbb{R}^q \times \mathbb{R}^q \rightarrow{\mathbb{R}}$ and $\kappa_Y: \mathbb{R}^p \times \mathbb{R}^p \rightarrow{\mathbb{R}}$ be Mercer kernels. $d_{\kappa_X}(\cdot,\cdot)$ and $d_{\kappa_Y}(\cdot,\cdot)$ are distance functions induced by $\kappa_X$ and $\kappa_Y$, respectively. Both distance functions have a bounded range $[0,1)$. For every $\epsilon > 0$,
\begin{align*}
    &\mathrm{Pr}\left[\mathrm{dCov}_{\kappa_X,\kappa_Y}^n - \mathrm{dCov}_{\kappa_X,\kappa_Y}(X,Y) \ge \epsilon \right] \le \exp{\left(\frac{-n\epsilon^2}{512}\right)},\\
    &\mathrm{and} \\
    &\mathrm{Pr}\left[\mathrm{dCov}_{\kappa_X,\kappa_Y}(X,Y) - \mathrm{dCov}_{\kappa_X,\kappa_Y}^n \ge \epsilon \right] \le \exp{\left(\frac{-n\epsilon^2}{512}\right)}.
\end{align*}
\end{thm}
The proof is provided in Appendix~\ref{app_c}. Note that the above result is established for both $X$ and $Y$ being numerical. When $Y$ is categorical, it can be embedded into $\mathbb{R}^K$ using the set difference kernel (\ref{setdiff}). Therefore, in the following discussion, we use the simpler notation introduced in Lemma~\ref{lem_dcovk} where $\mathrm{dCov}_{\kappa_X,\kappa_Y}$ is denoted by $\mathrm{dCov}_{\kappa_X}$.

The upper bounds for generalized Gini distance covariance is clearly tighter than those for generalized distance covariance. Replacing $\mathrm{gCov}_{\kappa}(X,Y)$ in $H_0$ and $H_1$ with $\mathrm{dCov}_{\kappa_X}(X,Y)$, one may develop dependence tests parallel to those in Section~\ref{ucb}: reject the null hypothesis when $\mathrm{dCov}_{\kappa_X}^n \ge c n^{-t}$ where $c > 0$ and $t\in \left(0, \frac{1}{2} \right)$. Upper bounds on Type I and Type II errors can be established in a result similar to Corollary~\ref{upperbound_gini} with the only difference being replacing the constant $12.5$ with $512$. Hence the bounds on the generalized Gini distance covariance based dependence test are tighter than those on the generalized distance covariance based dependence test.

To further compare the two dependence tests, we consider the following null and alternative hypotheses: 
\begin{eqnarray*}
    {H_0} & : & S(X,Y) = 0, \\
    {H_1} & : & S(X,Y) \ge \mathcal{T}, \; \mathcal{T} > 0,
\end{eqnarray*}
where $S(X,Y) =  \mathrm{gCov}_{\kappa_X}(X,Y)$ or $\mathrm{dCov}_{\kappa_X}(X,Y)$ with the corresponding test statistics $S_n =  \mathrm{gCov}_{\kappa_X}^n$ or $\mathrm{dCov}_{\kappa_X}^n$, respectively.
The null hypothesis is rejected when $S_n \ge \tau$ where $0 < \tau \le \mathcal{T}$. Note that this test is more general than the dependence test discussed in Section~\ref{ucb}, which is a special case with $\mathcal{T} = 2c n^{-t}$ and $\tau = c n^{-t}$. Upper bounds on Type I errors follow immediately from (\ref{gCovT1}) by replacing $c n^{-t}$ with $\tau$. Type II error bounds, however, are more difficult to derive due to the fact that $\tau = \mathcal{T}$ would make deviation nonexistent. Next, we take a different approach by establishing which one of $\mathrm{gCov}_{\kappa_X}^n$ and $\mathrm{dCov}_{\kappa_X}^n$ is less likely to underperform in terms of Type II errors.

Under the alternative hypothesis
\begin{eqnarray*}
    {H_1}' & : & \mathrm{dCov}_{\kappa_X}(X,Y) \ge \mathcal{T},\; \mathcal{T} > 0,
\end{eqnarray*}
we compare two dependence tests: 
\begin{itemize}
    \item accepting ${H_1}'$ when $\mathrm{gCov}_{\kappa_X}^n \ge \tau$, $0 < \tau \le \mathcal{T}$;
    \item accepting ${H_1}'$ when $\mathrm{dCov}_{\kappa_X}^n \ge \tau$, $0 < \tau \le \mathcal{T}$.
\end{itemize}  
We call that {\bf``$\mathrm{gCov}_{\kappa_X}^n$ underperforms} $\mathrm{dCov}_{\kappa_X}^n$" if and only if 
$$
\mathrm{gCov}_{\kappa_X}^n < \tau \le \mathrm{dCov}_{\kappa_X}^n,
$$
i.e., the dependence between $X$ and $Y$ is detected by $\mathrm{dCov}_{\kappa_X}^n$ but not by $\mathrm{gCov}_{\kappa_X}^n$. The following theorem demonstrates an upper bound on the probability that $\mathrm{gCov}_{\kappa_X}^n$ underperforms $\mathrm{dCov}_{\kappa_X}^n$.
\begin{thm}
Under ${H_1}'$ and conditions of Theorem~\ref{ucb_dCovk}, there exists $\gamma > 0$ such that the following inequality holds for any $\mathcal{T}>0$ and $0 < \tau \le \mathcal{T}$:
\begin{align*}
    &\mathrm{Pr}\left[ \mathrm{gCov}_{\kappa_X}^n\; \mathrm{underperforms}\; \mathrm{dCov}_{\kappa_X}^n | {H_1}' \right] \\
    &\le \exp{\left(\frac{-n\gamma^2}{12.5}\right)} + \exp{\left(\frac{-n\gamma^2}{512}\right)}.
\end{align*}
\end{thm}
\begin{proof}
Lemma~\ref{lem_dcovk} implies that $\mathrm{gCov}_{\kappa_X}(X,Y) \ge \mathrm{dCov}_{\kappa_X}(X,Y)$
where the equality holds if and only if both are $0$, i.e., $X$ and $Y$ are independent. Therefore, under
${H_1}'$, for any $\mathcal{T} > 0$ and $0 < \tau \le \mathcal{T}$, we define
$$
\gamma = \frac{\mathrm{gCov}_{\kappa_X}(X,Y) - \mathrm{dCov}_{\kappa_X}(X,Y)}{2} > 0.
$$
It follows that
\begin{align*}
    &\mathrm{Pr}\left[ \mathrm{gCov}_{\kappa_X}^n\; \mathrm{underperforms}\; \mathrm{dCov}_{\kappa_X}^n | {H_1}' \right] \\
    & = \mathrm{Pr}\left[ \mathrm{gCov}_{\kappa_X}^n < \tau \le \mathrm{dCov}_{\kappa_X}^n | {H_1}' \right]\\
    & \le \mathrm{Pr}\left[ \mathrm{gCov}_{\kappa_X}^n <  \mathrm{dCov}_{\kappa_X}^n | {H_1}' \right] \\
    & \le \mathrm{Pr}\left[ \mathrm{gCov}_{\kappa_X}(X,Y) - \mathrm{gCov}_{\kappa_X}^n  \ge \gamma \;\; \mathrm{OR} \right. \\
    & \;\;\;\;\;\left. \mathrm{dCov}_{\kappa_X}^n - \mathrm{dCov}_{\kappa_X}(X,Y) \ge \gamma | {H_1}'\right]\\
    & \le \mathrm{Pr}\left[ \mathrm{gCov}_{\kappa_X}(X,Y) - \mathrm{gCov}_{\kappa_X}^n  \ge \gamma | {H_1}' \right] \\
    & +\mathrm{Pr}\left[ \mathrm{dCov}_{\kappa_X}^n - \mathrm{dCov}_{\kappa_X}(X,Y) \ge \gamma | {H_1}'\right]\\
    & \le \exp{\left(\frac{-n\gamma^2}{12.5}\right)} + \exp{\left(\frac{-n\gamma^2}{512}\right)}
\end{align*}
where the last step is from Theorem~\ref{ucb_gCovk} and Theorem~\ref{ucb_dCovk}.
\end{proof}

\subsection{Asymptotic Analysis}\label{ab}
We now present asymptotic distributions for the proposed Gini covariance and Gini correlation.   
\begin{thm}
Assume $\E (d_{\kappa}^2(X, X^\prime))<\infty$ and $p_k>0$ for $k=1,...,K$. Under dependence of $X$ and $Y$, $\mathrm{gCov}_{\kappa_X}^n$ and $\mathrm{gCor}_{\kappa_X}^n$ have the asymptotic normality property. That is, 
\begin{eqnarray}
    \sqrt{n}(\mathrm{gCov}_{\kappa_X}^n-\mathrm{gCov}_{\kappa_X}(X,Y)) \xrightarrow{D} {\cal N}(0,\sigma_{v}^2), \label{an_cov}\\
    \sqrt{n}(\mathrm{gCor}_{\kappa_X}^n-\mathrm{gCor_{\kappa_X}}(X,Y)) \xrightarrow{D} {\cal N}(0,\frac{\sigma_{v}^2}{\Delta^2}), \label{an_cor}
\end{eqnarray}
where $\sigma_v^2$ is given in the proof. 

Under independence of $X$ and $Y$, $\mathrm{gCov}_{\kappa_X}^n$ and $\mathrm{gCor}_{\kappa_X}^n$ converge in distribution, respectively,  according to
\begin{eqnarray}
    n (\mathrm{gCov}_{\kappa_X}^n) \xrightarrow{D} \sum_{l=1}^\infty \lambda_l (\chi^2_{1l} -1), \label{chisq_cov}\\
    n (\mathrm{gCor}_{\kappa_X}^n) \xrightarrow{D} \frac{1}{\Delta}\sum_{l=1}^\infty \lambda_l (\chi^2_{1l} -1),\label{chisq_cor}
\end{eqnarray}
where $\lambda_1,...$ are non-negative constants dependent on $F$ and  $\chi^2_{11}, \chi^2_{12}, ...,$ are independent $\chi^2_1$ variates. 
\end{thm}
Note that 
the boundedness of the positive definite kernel $\kappa$ implies the condition of $\E (d_{\kappa}^2(X, X^\prime))<\infty$.  

\begin{proof}
 We focus on a proof for the generalized Gini distance covariance and results for the correlation follow immediately from Slutsky's theorem \cite{slutsky25} and the fact that $\hat{\Delta}$ is a consistent estimator of $\Delta$.  
 
  Let $g(x) =\E d_{\kappa}(x, X^\prime) - \E d_{\kappa}(X, X^\prime)$. With the U-statistic theorem, we have
$$\sqrt{n} (\hat{\Delta}-\Delta) \xrightarrow{D} N(0, v^2),$$
where $v^2 = 4 \E g^2(X) = 4\sum_k p_k \E g^2(X_k)$.  Similarly,
let $g_k(x) =\E d_{\kappa}(x, X_k^\prime) - \E d_{\kappa}( X_k, X_k^\prime)$  for $k=1,2,..., K$ and  $w_k^2 = 4 \E g_k(X_k)^2$. We have 
$$\sqrt{n_k} (\hat{\Delta}_k-\Delta_k) \xrightarrow{D} N(0, w_k^2), \;\;\; \mbox{ for } k=1,2,..., K.$$
Since $n_k = n \hat{p}_k \rightarrow n p_k $ in probability, by Slutsky's theorem we are able to write this result as
$$\sqrt{n} (\hat{\Delta}_k-\Delta_k) \xrightarrow{D} N(0, v_k^2),\;\;\;\; \mbox{ where }\;\;v_k^2 = w_k^2/p_k.$$

Let $\bi \Sigma $ be the variance and covariance matrix for $\tilde{\bi g} = 2 (g_1(X_1),...,g_K(X_K),g(X))^T$, where $X=X_k$ with probability $p_k$. In other words, $\bi \Sigma = \E \tilde{\bi g}\tilde{ \bi g}^T$. Denote $ (\hat{\Delta}_1, ..., \hat{\Delta}_K, \hat{\Delta})^T$ as $\hat{\bi \delta}$ and $( \Delta_1,...,\Delta_K, \Delta)^T$ as $\bi \delta $.  From the U-statistic theorem \cite{hoeffding48}, we have    $\sqrt{n}(\hat{\bi \delta}-\bi \delta) \xrightarrow{D} N(\bi 0, \bi \Sigma)$.  
Obviously, $\bi \Sigma$ has diagonal elements $v_1^2,...,v_K^2, v^2$.  
Let $\bi b=(-p_1,...,-p_K,1)^T$ be the gradient vector of $\mathrm{gCov}_{\kappa_X}(X,Y)$ with respect to $\bi \delta$. Then $\sigma_v^2 = \bi b^T \bi \Sigma \bi b >0$
under the assumption of dependence of $X$ and $Y$, since 
\begin{align*}
 &h(x):= \bi b^T \tilde{\bi g}(x) =2 \sum_kp_k( g(x_k)-g_k(x_k))\\
 &=2\sum_k p_k (\E d_{\kappa}(x_k, X)-\E d_{\kappa}(x_k, X_k)) -2( \Delta-\sum_k p_k \Delta_k) \\
 &\neq 0 
\end{align*}  and $\sigma_v^2 = \sum_kp_k\E[h(X_k)^2]$. In this case, by the Delta method,  $\sqrt{n}\bi b^T (\hat{\bi \delta}-\bi \delta)$ is asymptotically normally distributed with 0 mean and variance $\sigma_v^2$.  With the result of $\hat{\bi b} =(-\hat{p}_1,...,-\hat{p}_K,1)^T$ being a consistent estimator of $\bi b$ and by the Slutsky's theorem, we have the same limiting normal distribution for $\mathrm{gCov}_{\kappa_X}^n = \hat{\bi b}^T\hat{\bi \delta}$ as that of $\bi b^T \hat{\bi \delta}$.   Therefore, the result of (\ref{an_cov}) is proved.  

However, under the independence assumption, $\sigma_v^2 =0$ because 
\begin{align*}
    &h(x) \\
    &=2\sum_k p_k (\E d_{\kappa}(x_k, X)-\E d_{\kappa}(x_k, X_k)) -2( \Delta-\sum_k p_k \Delta_k) \\
    &\equiv 0,
\end{align*}
resulting from the same distribution of $X$ and $X_k$. This corresponds to the degenerate case of U-statistics and  $\bi b^T \hat{\bi \delta}$ has a mixture of $\chi^2$ distributions \cite{serfling80}. 
Hence  the result of (\ref{chisq_cov}) holds.  
\end{proof}

One way to use the results of (\ref{an_cov}) and (\ref{an_cor}) is to test $H_0$ based on the confidence interval approach. More specifically,  an asymptotically $(1-\alpha)100\%$ confidence interval for $\mathrm{gCov}_{\kappa_X}(X,Y)$ is 
$$\mathrm{gCov}_{\kappa_X}^n(X,Y) \pm Z_{1-\alpha/2} \frac{\hat{\sigma}_v^2}{\sqrt{n}},$$
where $\hat{\sigma}_v^2$ is a consistent estimator of $\sigma_v^2$ and $Z_{1-\alpha/2}$ is the $1-\alpha/2$ quantile of the  standard normal random variable.   If this interval does not contain 0, we can reject $H_0$ at significance level $\alpha/2$. This test controls Type II error to be $\alpha/2$. 

On the other hand, if a test to control Type I error is preferred, we usually need to rely on a permutation test rather than the results of (\ref{chisq_cov}) and (\ref{chisq_cor}) since $\lambda$'s depend on the distribution $F$, which is unknown. Details of the permutation test are in the next section.   

\section{An Algorithmic View}\label{discussions}
Although the uniform convergence bounds for generalized Gini distance covariance and generalized Gini distance correlation in Section~\ref{ucb} and Section~\ref{connections} are established upon the bounded kernel assumption, all the results also hold for Gini distance covariance and Gini distance correlation if the features are bounded. This is because when the features are bounded, they can be normalized so that $\sup_{x,x'}|x-x'|_q = 1$.

The calculation of test statistics (\ref{gCovEst}) and (\ref{gCorEst}) requires evaluating distances between all unique pairs of samples. Its time complexity is therefore $\Theta(n^2)$, where $n$ is the sample size. In the one dimension case, i.e., $q=1$, Gini distance statistics can be calculated in $\Theta(n\log n)$ time~\cite{dang18}~\footnote{When the inner produce kernel $\kappa(x,x') = x^Tx'$ is chosen, generalized Gini distance statistics reduces to Gini distance statistics.}. Note that distance covariance and distance correlation can also be calculated in $\Theta(n\log n)$ time~\cite{hou16}. Nevertheless, the implementation for Gini distance statistics is much simpler as it does not require the centering process.

\begin{figure*}[t]
\hskip -.2in
\centering
\begin{tabular}{cc}
    \includegraphics[draft=false, width = 3.5in, keepaspectratio]{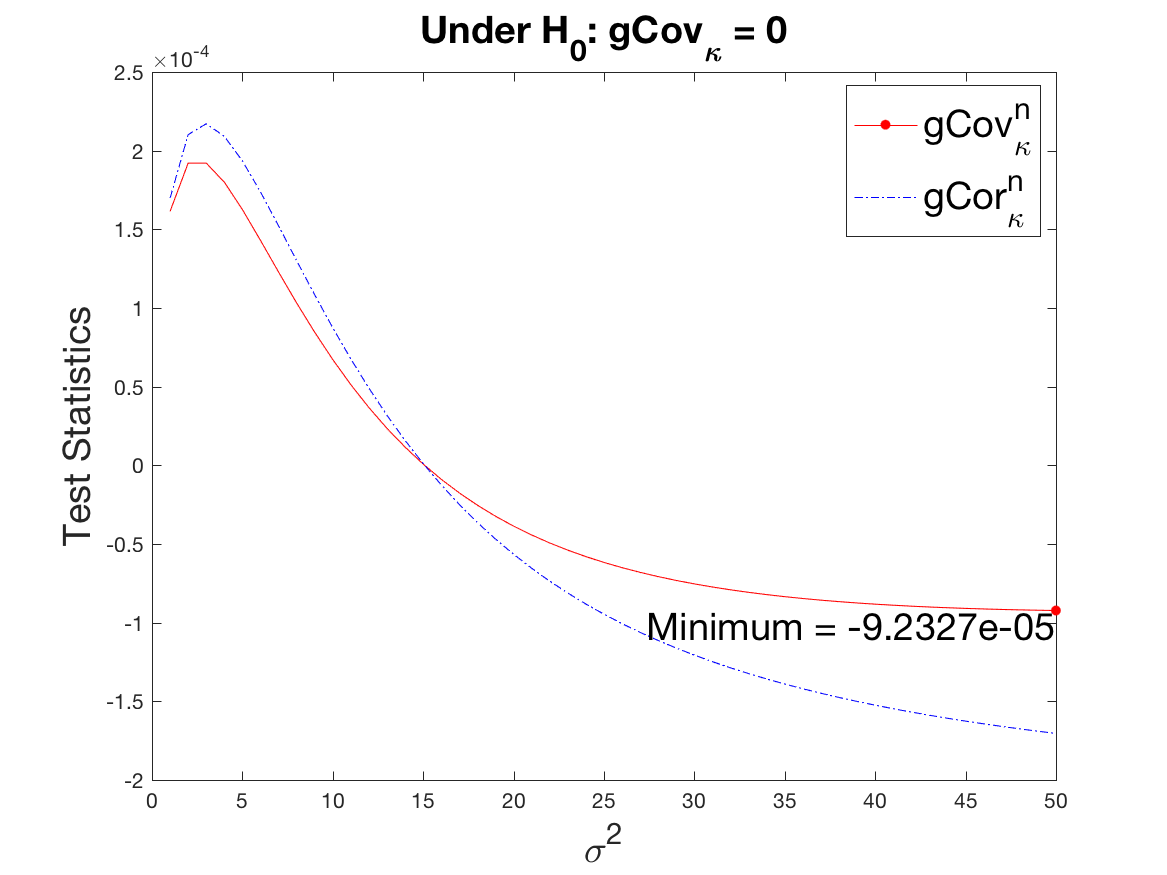}
    & \hskip -.4in \includegraphics[draft=false, width = 3.5in, keepaspectratio]{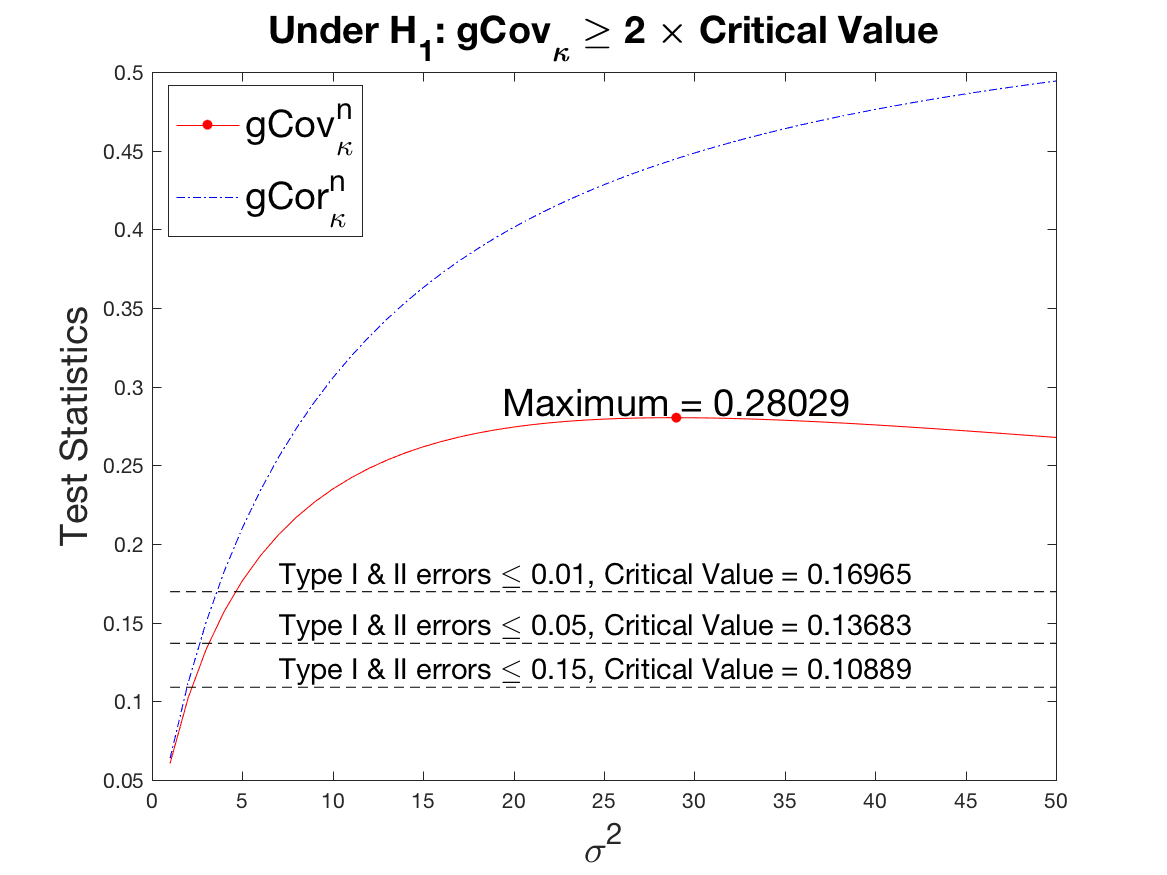} \\
    (a) & \hskip -.4in (b)
\end{tabular}
\caption{Estimates of the generalized Gini distance covariance and generalized Gini distance correlation for different kernel parameters using $2000$ iid samples: (a) independent case; (b) dependent case. Three critical values are shown in (b). They are calculated for significance levels $0.01$, $0.05$, and $0.15$, respectively. In terms of the uniform convergence bounds, the optimal value of the kernel parameter $\sigma$ is defined by the minimizer (or maximizer) of the test statistics under $H_0$ (or $H_1$). } \label{figure_sigma}
\end{figure*}

Generalized Gini distance statistics are functions of the kernel parameter $\sigma$. Figure~\ref{figure_sigma}(a) shows $\mathrm{gCov}_\kappa^n$ and $\mathrm{gCor}_\kappa^n$ of $X_1$ and $Y_1$ for $n=2000$. The numerical random variable $X_1$ is generate from a mixture of two dimensional normal distributions: $N_1 \sim \mathcal{N}\left([1,2]^T,\mathrm{diag[2,.5]}\right)$,$N_2 \sim \mathcal{N}\left([-3,-5]^T,\mathrm{diag[1,1]}\right)$, and $N_3 \sim \mathcal{N}\left([-1,2]^T,\mathrm{diag[2,2]}\right)$.
The three components have equal mixing proportions. The categorical variable $Y_1 \in \{y_1,y_2,y_3\}$ is independent of $X_1$. The results in Figure~\ref{figure_sigma}(b) are calculated from $X_2$ and $Y_2$ for $n=2000$. The numerical random variable $X_2$ is generated by $N_i$ if and only if $Y_2 = y_i$, $i=1,2,3$. The categorical distribution of $Y_2$ is $\mathrm{Pr}(Y_2 = y_i) = \frac{1}{3}$. It is clear that $X_2$ and $Y_2$ are dependent on each other. Figure~\ref{figure_sigma} shows the impact of kernel parameter $\sigma$ on the estimated generalized Gini distance covariance and Gini distance correlation. As a result, this affects the Type I and Type II error bounds given in Section~\ref{ucb}. In this example, under $H_0$ (or $H_1$), the minimum (or maximum) $\mathrm{gCov}_\kappa^n$ is achieved at $\sigma^2 = 50$ (or $\sigma^2 = 29$). These extremes yield tightest bounds in (\ref{gCovT1}) and (\ref{gCovT2})~\footnote{The kernel parameter $\sigma$ also affects the Type I and Type II error bounds for $\mathrm{gCor}_\kappa^n$ in (\ref{gCorT1}) and (\ref{gCorT2}). The Type I error bound for $\mathrm{gCov}_\kappa^n$ is significantly tighter than that for $\mathrm{gCor}_\kappa^n$.}. Note that $\mathrm{gCov}_\kappa^n$ is an unbiased estimate of $\mathrm{gCov}_\kappa$. Although $\mathrm{gCov}_\kappa$ can never be negative, $\mathrm{gCov}_\kappa^n$ can be negative, especially under $H_0$. 

This example also suggests that in addition to the theoretical importance, the inequalities in (\ref{gCovT1}) and (\ref{gCovT2}) may be directly applied to dependence tests. Given a desired bound (or significance level), $\alpha$, on Type I and Type II errors, we call the value that determines whether $H_0$ should be rejected (hence to accept $H_1$) the \emph{critical value} of the test statistic. Based on (\ref{gCovT1}) and (\ref{gCovT2}), the critical value for $\mathrm{gCov_\kappa^n}$, $\mathrm{cv}(\alpha,n)$, which is a function of $\alpha$ and the sample size $n$, is calculated as
$$
\mathrm{cv}(\alpha,n) = \sqrt{\frac{12.5 \log \frac{1}{\alpha}}{n}}.
$$
The three horizontal dashed lines in Figure~\ref{figure_sigma}(b) illustrate the critical values for $\alpha=0.01$, $\alpha=0.05$, and $\alpha=0.15$, respectively. The population Gini distance covariance estimated using $20,000$ iid samples is not included in the figure because of its closeness to $\mathrm{gCov_\kappa^n}$. With a proper choice of $\sigma$, $H_1$ should be accepted based on the $2000$ samples of $(
X_2, Y_2)$ with both Type I and Type II errors no greater than $0.05$. Note that we could not really accept $H_1$ at the level $\alpha=0.01$ because the estimated maximum $\mathrm{gCov}_\kappa$ is around $0.28$, which is smaller than $0.3393$ (two times the critical value at $\alpha=0.01$).

\begin{figure*}[t]
\hskip -.2in
\centering
\begin{tabular}{cc}
    \includegraphics[draft=false, width = 3.5in, keepaspectratio]{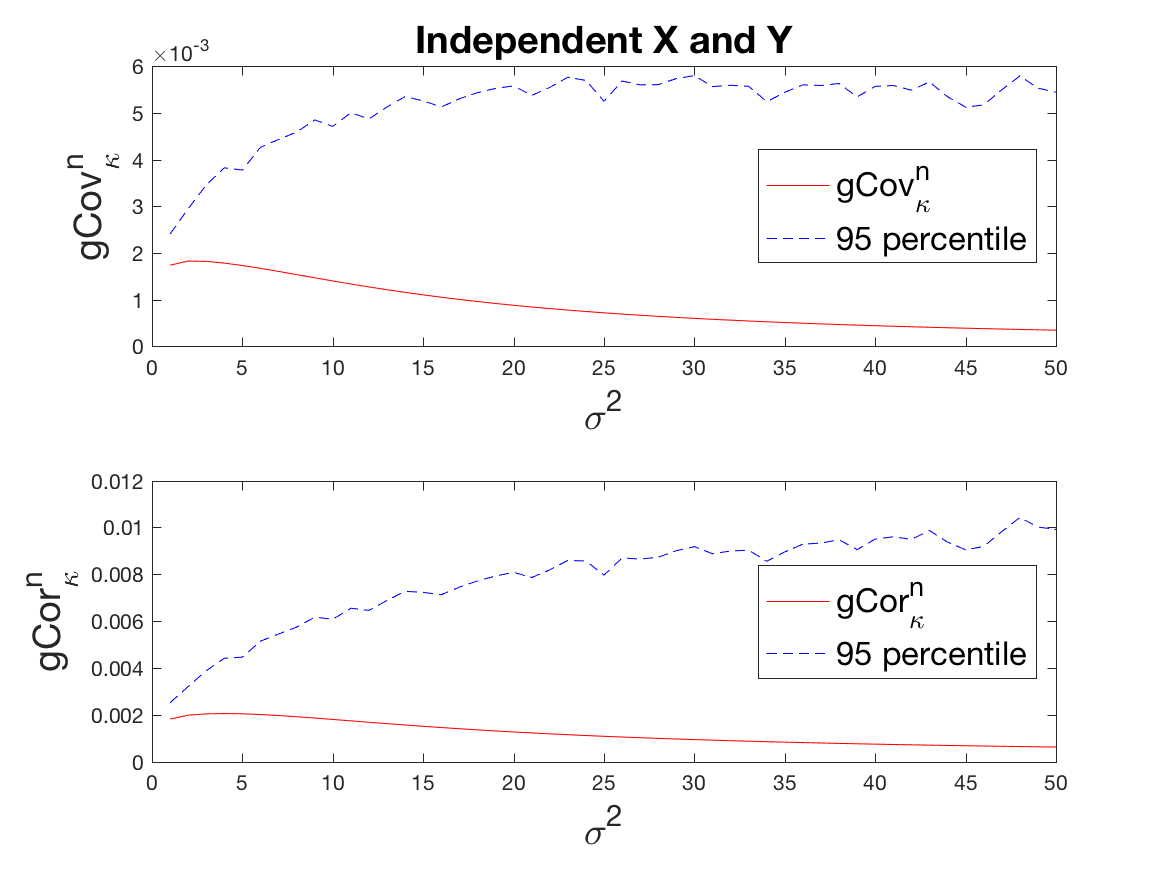}
    & \hskip -.4in \includegraphics[draft=false, width = 3.5in, keepaspectratio]{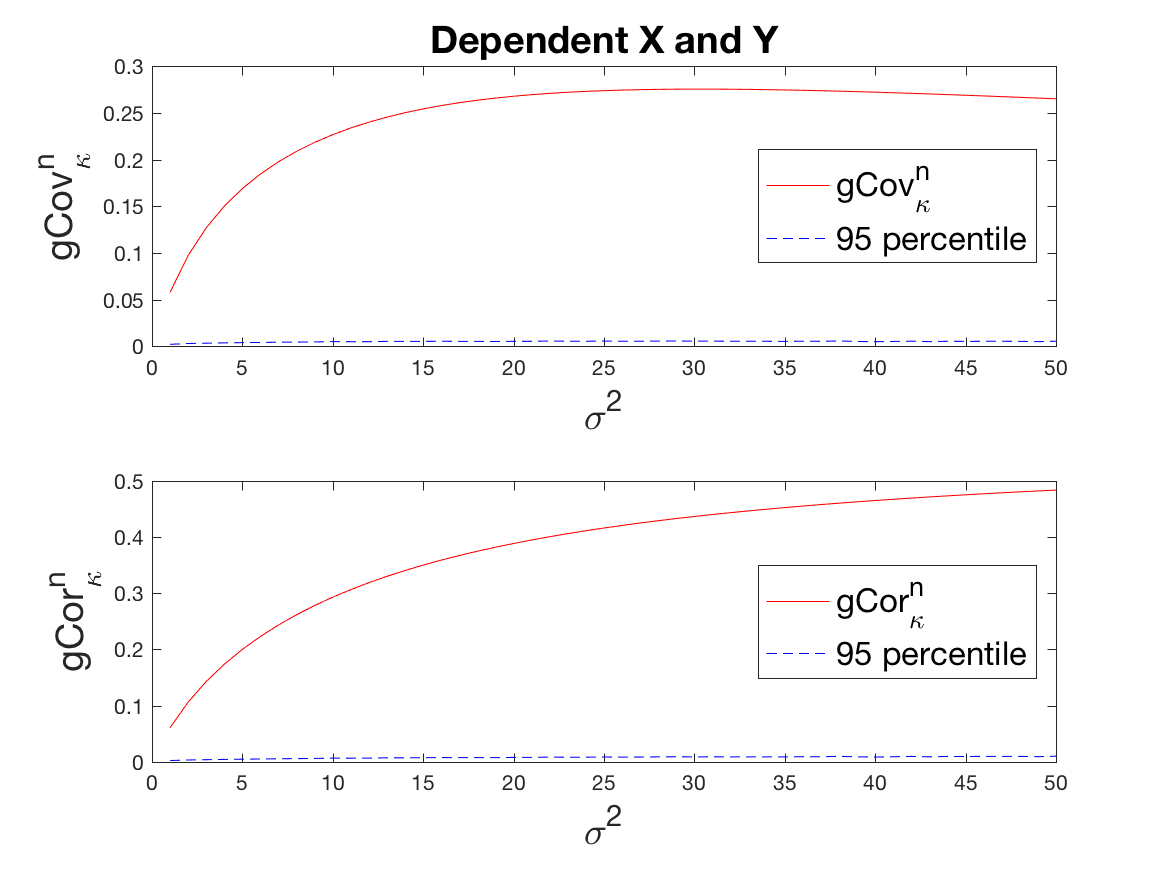} \\
    (a) & \hskip -.4in (b)
\end{tabular}
\caption{Permutation tests of the generalized Gini distance covariance and generalized Gini distance correlation for different kernel parameters using $200$ iid samples and $5000$ random permutations: (a) independent case; (b) dependent case. The $95$ percentile curves define the critical values for $\alpha = 0.05$. They are calculated from the permuted data. Test statistics higher (lower) than the critical value suggest accepting $H_1$ ($H_0$).} \label{figure_perm}
\end{figure*}
The above test, although simple, has two limitations:
\begin{itemize}
    \item Choosing an optimal $\sigma$ is still an open problem. Numerical search is computationally expensive even if it is in one dimension;
    \item The simplicity of the distribution free critical value $\mathrm{cv}(\alpha,n)$ comes with a price: it might not be tight enough for many distributions, especially when $n$ is small.
\end{itemize}
Therefore, we apply permutation test~\cite{edgington07}, a commonly used statistical tool for constructing sampling distributions, to handle scenarios that the test based on $\mathrm{cv}(\alpha,n)$ is not feasible. We randomly shuffle the samples of $X$ and keep the samples of $Y$ untouched. We expect the generalized Gini distance statistics of the shuffled data should have values close to $0$ because the random permutation breaks the dependence between samples of $X$ and samples of $Y$. Repeating the random permutation many times, we may estimate the critical value for a given significance level $\alpha$ based on the statistics of the permuted data. A simple approach is to use the percentile defined by $\alpha$, e.g., when $\alpha = 0.05$ the critical value is $95$-th percentile of the test statistic of the permuted data. The null hypothesis $H_0$ is rejected when the test statistic is larger than the critical value. 

Figure~\ref{figure_perm} shows the permutation test results of data generated from the same distributions used in Figure~\ref{figure_sigma}. The plots on the top are $\mathrm{gCov}_\kappa^n$ and the critical values. The plots at the bottom are $\mathrm{gCor}_\kappa^n$ and the critical values. Test statistics are calculated from $500$ samples. The critical values are estimated from $5000$ random permutations at $\alpha = 0.05$. As illustrated in Figure~\ref{figure_perm}(a), when $X$ and $Y$ are independent, the permutation tests do not reject $H_0$ at significance level $0.05$. Figure~\ref{figure_perm}(b) shows that when $X$ and $Y$ are dependent, $H_0$ is rejected at significance level $0.05$ by the permutation tests. It is interesting to note that the decision to reject or accept $H_0$ is not influenced by the value of the kernel parameter $\sigma$.~\footnote{Although the decision to reject or accept $H_0$ is not affected by $\sigma$, the $p$-value of the test does vary with respect to $\sigma$.}

\section{Experiments}\label{experiments}
We first compare Gini distance statistics with distance statistics on artificial datasets where the dependent features are known. We then provide comparisons on the MNIST dataset, a breast cancer dataset and 15 UCI datasets. For these real datasets, we also include Pearson $R^2$ as a method under comparison.
\subsection{Simulation Results}
In this experiment, we compare dependence tests using four statistics, $\mathrm{dCov}_\kappa^n$, $\mathrm{dCor}_\kappa^n$, $\mathrm{gCov}_\kappa^n$, and $\mathrm{gCor}_\kappa^n$, on artificial datasets. The data were generated from three distribution families: normal, exponential, and Gamma distributions under both $H_0$ ($X$ and $Y$ are independent) and $H_1$ ($X$ and $Y$ are dependent). 
Given a distribution family, we first randomly choose a distribution $F_0$ and generate $n$ iid samples of $X$. Samples of the categorical $Y$ are then produced independent of $X$. 
Repeating the process, we create a total of $m$ independent datasets under $H_0$. The test statistics of these $m$ independent datasets are used in calculating the critical values for different Type I errors. In the dependent case, $X$ is produced by $F = \sum_{k=1}^K p_k F_k$, a mixture of $K$ distributions where $K$ is the number of different values that $Y$ takes, $p_k$ is the probability that $Y=y_k$, and $F_k$ is a distribution from the same family that yields the data under $H_0$. The dependence between $X$ and $Y$ is established by the data generating process: a sample of $X$ is created by $F_k$ if $Y = y_k$. The mixture model is randomly generated, i.e., $p_k$ and $F_k$ are both randomly chosen. For each $Y=y_k$ ($k=1,\cdots,K$), $n_k = n \cdot p_k$ iid samples of $X$ are produced following $F_k$. This results in one data set of size $n = \sum_{k=1}^K n_k$ under $H_1$. Following the same procedure, we obtain $m$ independent data sets under $H_1$ each corresponding to a randomly selected $K$-component mixture model $F$. The power of a test statistic, defined as $1 - \mathrm{Type\; II\; error}$, is then estimated using these $m$ independent data sets under $H_1$ and the critical values computed from the $m$ data sets under $H_0$. In our experiments, $n=100$ and $m = 10,000$. Table~\ref{table_distributions} summarizes the model parameters of the three distribution families. $I(\cdot)$ is the indicator function. $\mathcal{N}(a,b)$ denotes the normal distribution with mean $a$ and standard deviation $b$. $\mathcal{U}(a,b)$ denotes the uniform distribution over interval $[a,b]$. A distribution ($F_0$ or $F_k$) is randomly chosen via its parameter(s). The unbiasedness of $\mathrm{gCov}_\kappa^n$ requires that there are at least two data points for each value of $Y$. Therefore, random proportions that do not meet this requirement are removed. 
\begin{table}[t]
\caption{Models of Different Distribution Families} \label{table_distributions}
\centering
\begin{tabular}{ ||l||c|c|| } 
 \hline\hline
 & $p(x|\theta)$ & $\theta$  \\ 
 \hline\hline
 Normal & $\frac{1}{\sqrt{2 \pi \sigma^2}} e^{-\frac{(x-\mu)^2}{\sigma^2}}$ & $\mu \sim \mathcal{N}(0,5)$, $\sigma \sim \mathcal{U}(0,5)$\\ 
 Exponential & $\lambda e^{-\lambda x} \mathrm{I}(x \ge 0)$ & $\lambda \sim \mathcal{U}(0,5)$  \\ 
 Gamma   & $\frac{\beta^\alpha x^{\alpha - 1}e^{-\beta x}}{\Gamma(\alpha)}\mathrm{I}(x\ge 0)$ & $\alpha \sim \mathcal{U}(0,10)$,$\beta \sim \mathcal{U}(0,10)$\\ \hline
 Proportions & \multicolumn{2}{c||}{ $p_k=\frac{u_k}{\sum_{k=1}^Ku_k}$, $u_k \sim \mathcal{U}(0,1)$}\\ 
 \hline\hline
\end{tabular}
\end{table}

Table~\ref{table_power_auc} illustrates the performance of the four test statistics under different values of $K$ for the three distribution families with a fixed kernel parameter ($\sigma^2=10$). Two measures are computed from ROC: $\mathrm{Power}(\alpha =0.05)$ and Area Under Curve (AUC), where $\mathrm{Power}(\alpha =0.05)$ is calculated from ROC at Type I error is $0.05$. Both measures have values between $0$ and $1$ with a value closer to $1$ indicating better performance. The highest power and AUC among the four test statistics are shown in bold and the second highest are underlined. In this experiment, $\mathrm{gCov}_\kappa^n$ appears to be the most competitive test statistics in terms of ROC related measures at all values of $K$. In addition, both $\mathrm{gCov}_\kappa^n$ and $\mathrm{gCor}_\kappa^n$ outperform $\mathrm{dCov}_\kappa^n$ and $\mathrm{dCor}_\kappa^n$ in most of the cases. We also tested the influence of $\sigma^2$ and observed stable results (figures are provided in supplementary materials).
\begin{table*}[t]
\caption{Power ($\alpha = 0.05$) and AUC.} \label{table_power_auc}
\centering
\begin{tabular}{||c|l|cccc|cccc||}
\hline\hline
\multicolumn{2}{||c|}{} & \multicolumn{4}{c|}{Power} & \multicolumn{4}{c||}{AUC}\\
\cline{3-10}
\multicolumn{2}{||c|}{}& $\mathrm{dCov}_\kappa^n$ & $\mathrm{dCor}_\kappa^n$ & $\mathrm{gCov}_\kappa^n$ & $\mathrm{gCor}_\kappa^n$& $\mathrm{dCov}_\kappa^n$ & $\mathrm{dCor}_\kappa^n$ & $\mathrm{gCov}_\kappa^n$ & $\mathrm{gCor}_\kappa^n$\\
\hline
\multirow{3}{*}{$K=3$} & Normal & 0.977& 0.976& \textbf{0.984}& \underline{0.979}& \underline{0.994}& 0.993& \textbf{0.995}& \underline{0.994} \\
 & Exponential & 0.709& 0.705& \textbf{0.730}& \underline{0.715}& 0.887& 0.890& \underline{0.894}& \textbf{0.895} \\
 & Gamma & 0.971& 0.972& \textbf{0.979}& \underline{0.976}& 0.991& \underline{0.992}& \textbf{0.993}& \textbf{0.993} \\
\hline
\multirow{3}{*}{$K=4$} & Normal & \underline{0.993}& 0.992& \textbf{0.995}& \underline{0.993}& \underline{0.998}& \underline{0.998}& \textbf{0.999}& \underline{0.998} \\
 & Exponential & 0.779& 0.769& \textbf{0.799}& \underline{0.781}& 0.922& 0.922& \textbf{0.928}& \underline{0.927} \\
 & Gamma & 0.992& 0.991& \textbf{0.994}& \underline{0.993}& \textbf{0.998}& \textbf{0.998}& \textbf{0.998}& \textbf{0.998} \\
\hline
\multirow{3}{*}{$K=5$} & Normal & 0.997& 0.996& \textbf{0.999}& \underline{0.998}& \underline{0.999}& \underline{0.999}& \textbf{1.000}& \underline{0.999} \\
 & Exponential & \underline{0.821}& 0.809& \textbf{0.839}& 0.818& 0.940& 0.939& \textbf{0.947}& \underline{0.944} \\
 & Gamma & \underline{0.997}& \underline{0.997}& \textbf{0.998}& \textbf{0.998}& \underline{0.999}& \underline{0.999}& \textbf{1.000}& \underline{0.999} \\
\hline\hline
\end{tabular}
\end{table*}

\subsection{The MNIST Dataset}
We first tested feature selection methods using different test statistics on the MNIST data. The advantage of using an image dataset like MNIST is that we can visualize the selected pixels. We expect useful/dependent pixels to appear in the center part of the image. Some descriptions of the MNIST data are listed in TABLE~\ref{table_UCI_summary}.

The 5 test statistics under comparisons are: Pearson $R^2$, $\mathrm{dCov}_\kappa^n$, $\mathrm{dCor}_\kappa^n$, $\mathrm{gCov}_\kappa^n$, and $\mathrm{gCor}_\kappa^n$. For each method, the top $k$ features were selected by ranking the test statistics in descending order. Then the selected feature set was used to train the same classifier and the test accuracies were compared. The classifier used was a random forest consisting of 100 trees. We used the training and test set provided by ~\cite{lecun-mnisthandwrittendigit-2010} for training and testing. However, due to the size of the training set, using all training samples to calculate Gini and distance statistics is too time-consuming. Therefore, we randomly selected 5000 samples from the training set to calculate test statistics for all methods. For Gini and distance statistics, each feature was standardized by subtracting the mean and dividing by the standard deviation. The kernel parameter $\sigma^2$ was set to be 10. Due to the randomness involved in training random forest, each experiment was repeated 10 times and the average test accuracy was used for comparison.

\begin{figure*}
\hskip -.2in
\centering
\begin{tabular}{cc}
    \includegraphics[draft=false, width = 3.2in, keepaspectratio]{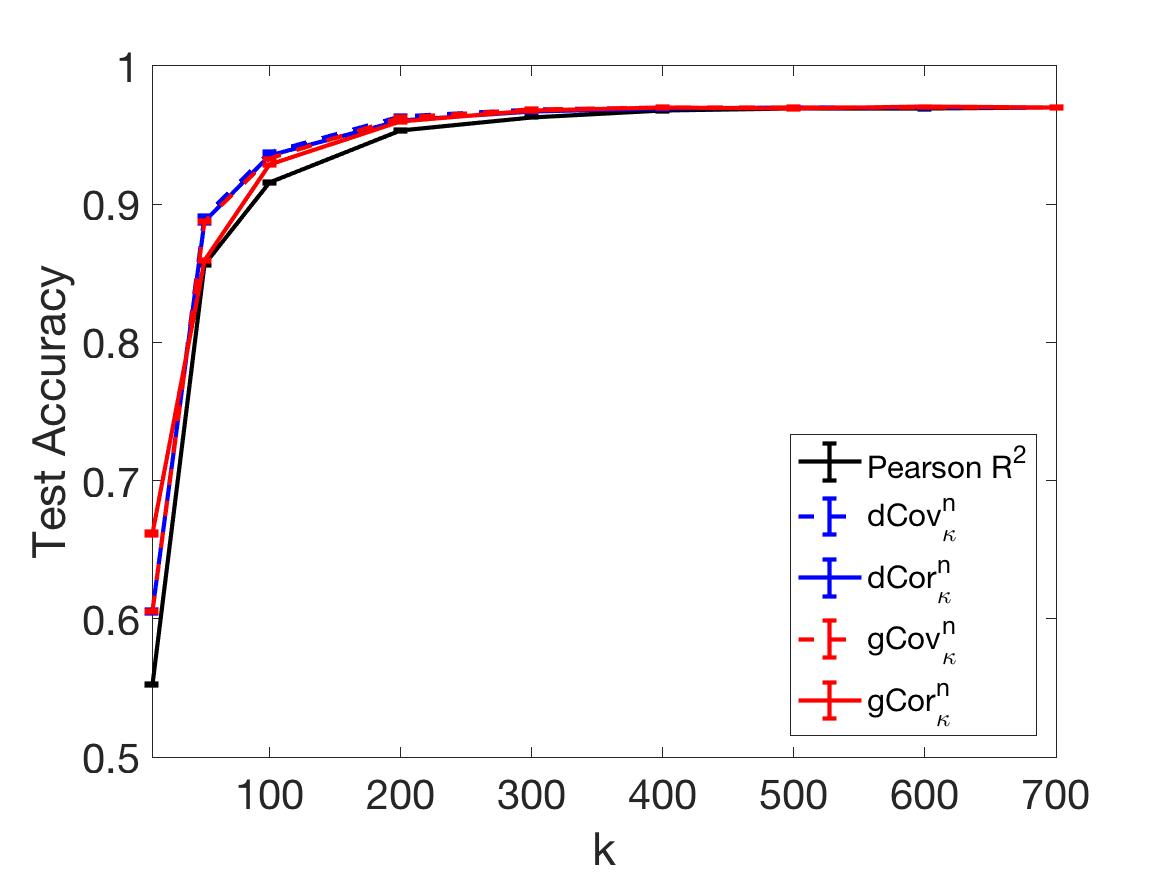} 
    & \hskip -.4in \includegraphics[draft=false, width = 3.2in, keepaspectratio]{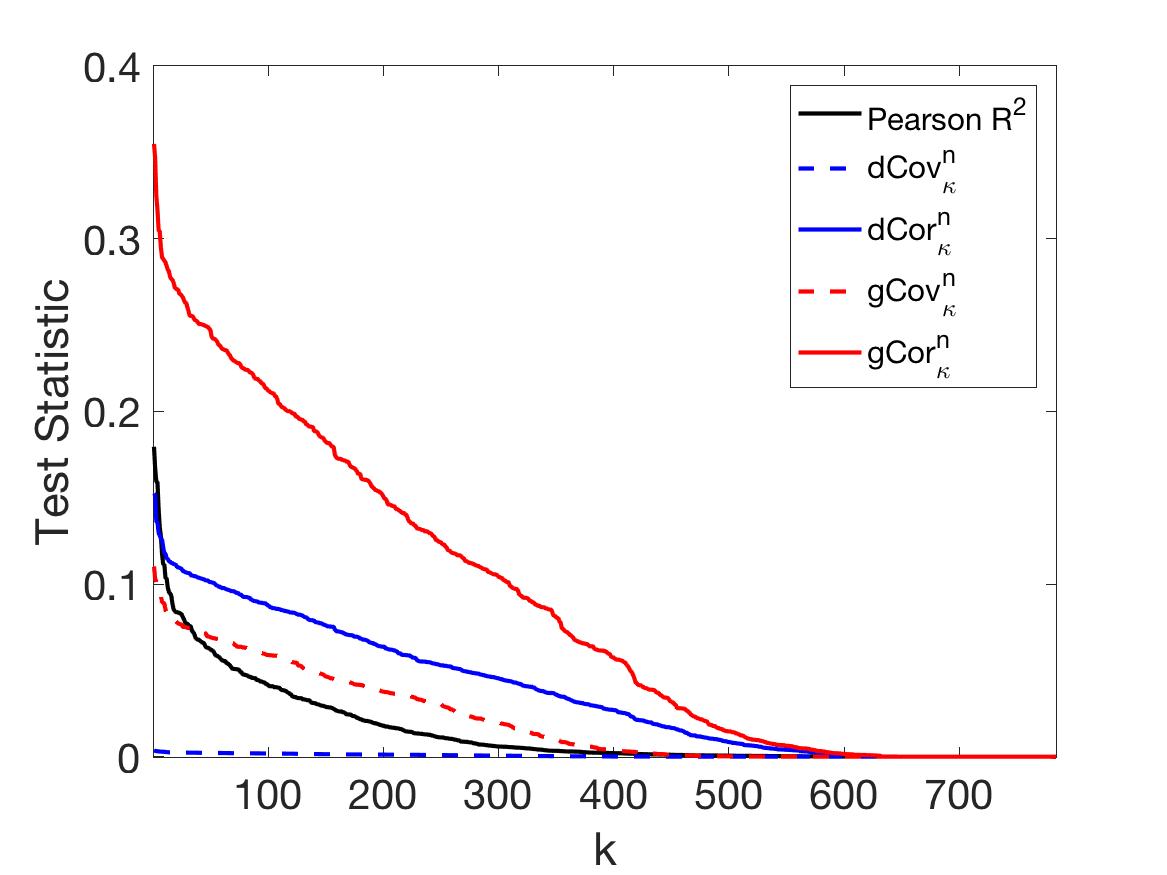} \\
    (a) & \hskip -.4in (b) \\
    \multicolumn{2}{c}{\includegraphics[draft=false,width = 6.25in, keepaspectratio]{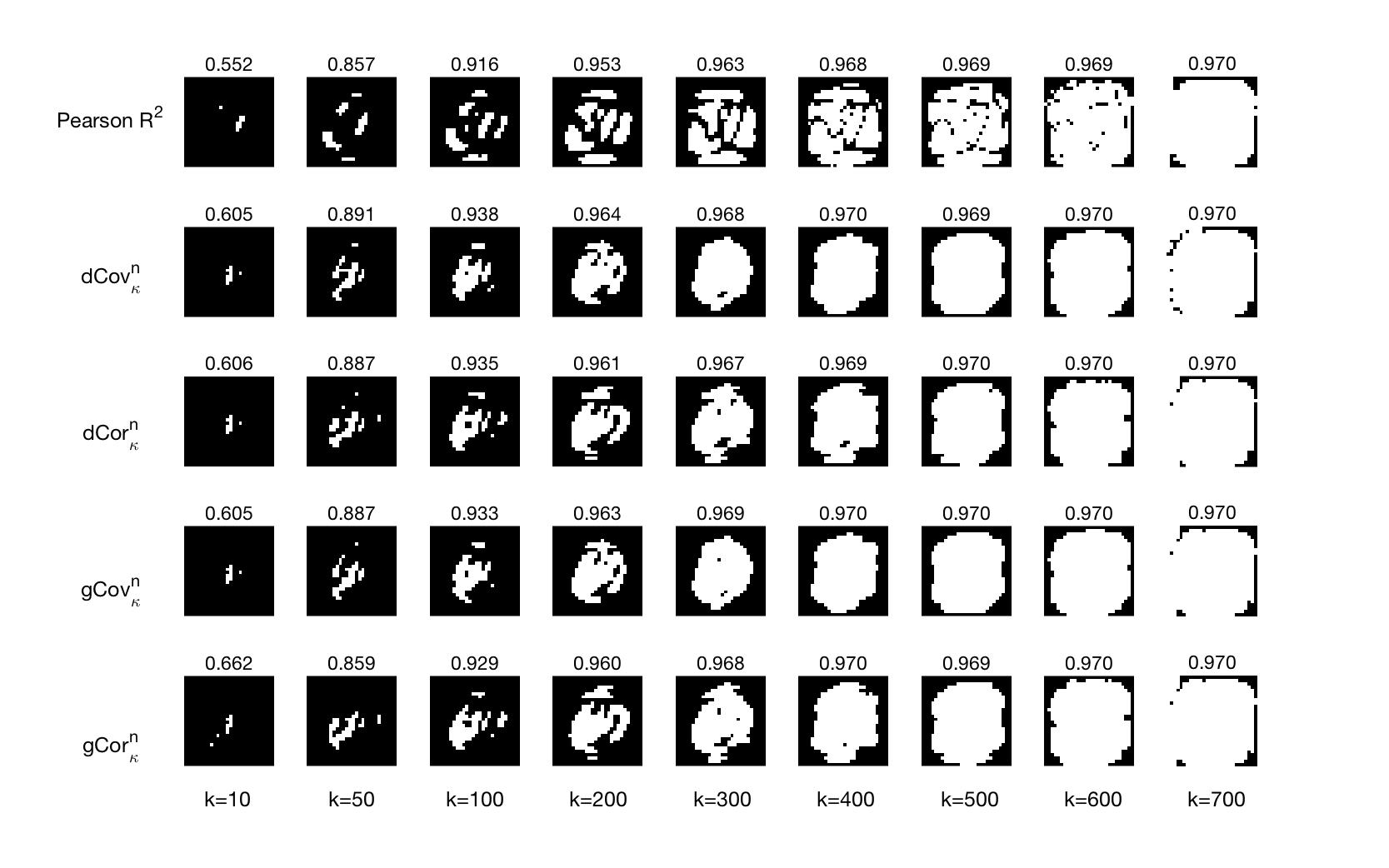}} \\
    \multicolumn{2}{c}{(c)}
\end{tabular}
\caption{The MNIST dataset. (a) Test accuracy using the top $k$ selected features. (b) Test statistics of features in descending order. (c) Visualization of the top $k$ pixels selected. White: selected. Black: not selected. Test accuracy using the selected pixels is labeled on the top of each image.} 
\label{figure:mnist} 
\end{figure*}

The results of the MNIST data are summarized in Fig.~\ref{figure:mnist}. Fig.~\ref{figure:mnist}(a) shows the test accuracies using the top $k$ features selected by different methods. Fig.~\ref{figure:mnist}(b) shows the test statistics in descending order. Fig.~\ref{figure:mnist}(c) shows the visualization of the selected pixels as white. From Fig.~\ref{figure:mnist}(a) we can see the clear increasing trend in accuracy as more features are selected, as expected. Among all methods, Pearson $R^2$ performs the poorest. This is because it is unable to select some of the pixels in the center part of the image as dependent features even with $k=600$, as shown in Fig.~\ref{figure:mnist}(c). The discrepancy in accuracy between Pearson $R^2$ and the other four test statistics comes from the ability of characterizing non-linear dependence. The other four methods behave similar except that $\mathrm{gCor}_\kappa^n$ has significant higher accuracy with $k=10$ and lower accuracy with $k=50$ than the other three ($\mathrm{dCov}_\kappa^n$, $\mathrm{dCor}_\kappa^n$, and $\mathrm{gCov}_\kappa^n$). Specifically, we expect $\mathrm{dCov}_\kappa^n$ and $\mathrm{gCov}_\kappa^n$ to behave very similarly because MNIST is a balanced dataset. Under the following two scenarios $\mathrm{dCov}_\kappa(X,Y)$ and $\mathrm{gCov}_\kappa(X,Y)$ will give the same ranking of the the features because their ratio is a constant (Remark 2.8 \& 2.9 of ~\cite{dang18}):
\begin{enumerate}
    \item When the data is balanced, i.e., $p_1=p_2=...=p_K=\frac{1}{K}$, $\mathrm{dCov}_{\kappa}(X,Y)=\frac{1}{K}\mathrm{gCov}_{\kappa}(X,Y)$;
    \item When the data has only 2 classes, i.e., $K=2$, $\mathrm{dCov}_{\kappa}(X,Y)=2p_1p_2\mathrm{gCov}_{\kappa}(X,Y)$.
\end{enumerate}
Hence, when $n$ is sufficiently large, $\mathrm{dCov}_\kappa^n$ and $\mathrm{gCov}_\kappa^n$ will have the same ranking for the features.

The difference between Gini and distance statistics is more observable in the value range, as shown in Fig.~\ref{figure:mnist}(b). Both $\mathrm{dCor}_\kappa^n$ and $\mathrm{gCor}_\kappa^n$ are bounded between 0 and 1, but clearly $\mathrm{gCor}_\kappa^n$ takes a much wider range than $\mathrm{dCor}_\kappa^n$. Therefore, $\mathrm{gCor}_\kappa^n$ is a more sensitive measure of dependence than $\mathrm{dCor}_\kappa^n$. $\mathrm{gCov}_\kappa^n$ is also more sensitive than $\mathrm{dCov}_\kappa^n$ as shown both empirically in Fig.~\ref{figure:mnist}(b) and theoretically by (\ref{gCov>=dCov}).

\subsection{The Breast Cancer Dataset}
We then compared the 5 feature selection methods on a gene selection task. The dataset used in this experiment was the TCGA breast cancer microarray dataset from the UCSC Xena database~\cite{Goldman326470}. This data contains expression levels of 17278 genes from 506 patients and each patient has a breast cancer subtype label (luminal A, luminal B, HER2-enriched, or basal-like). PAM50 is a gene signature consisting of 50 genes derived from microarray data and is considered as the gold-standard for breast cancer subtype prognosis and prediction~\cite{parker2009supervised}. In this experiment, we randomly hold-out 20\% as test data, used each method to select top $k$ genes, evaluate the classification performance and compare the selected genes with the PAM50 gene signature. Because this dataset has a relatively small sample size, all training examples were used to calculate test statistics and train the classifier, and we repeated each classification test 30 times. Other experiment setups were kept the same as previous.

\begin{figure*}
\hskip -.2in
\centering
\begin{tabular}{cc}
    \includegraphics[draft=false, width = 3.5in, keepaspectratio]{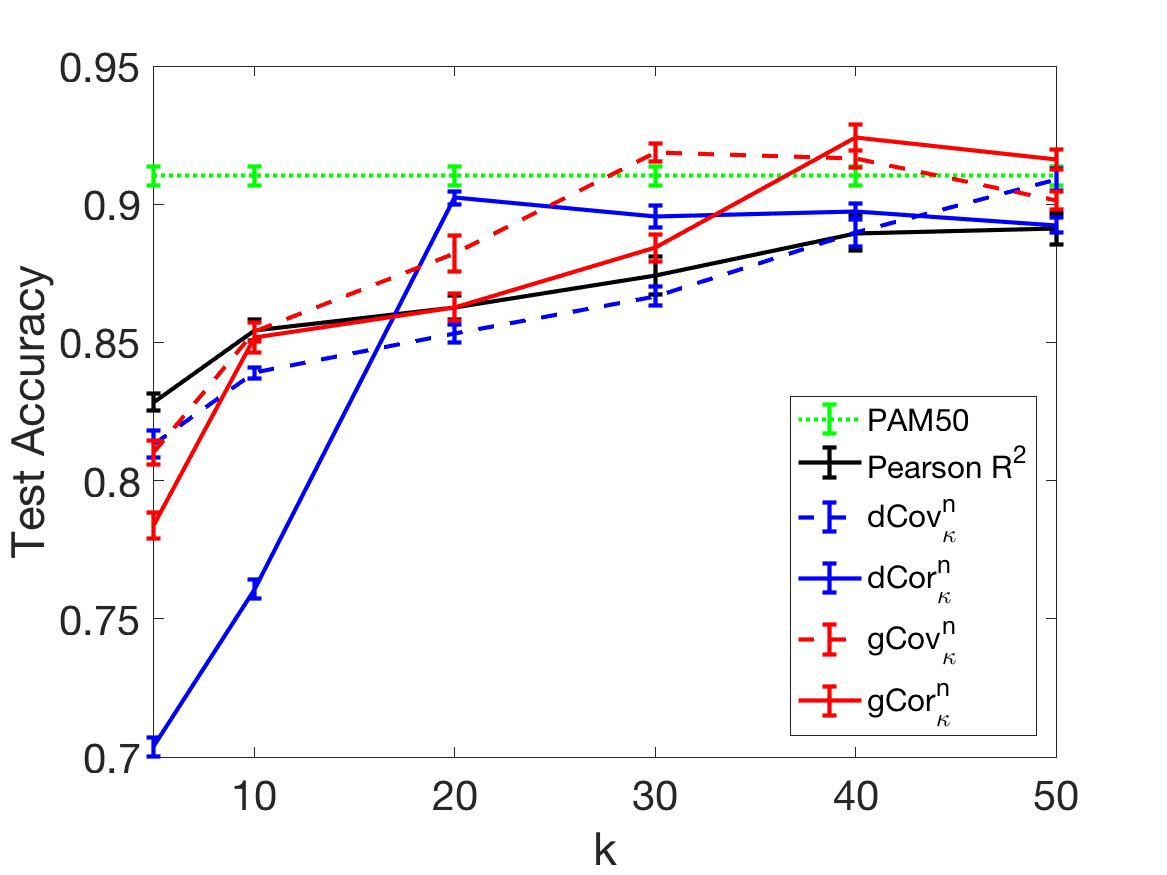} 
    & \hskip -.4in \includegraphics[draft=false, width = 3.5in, keepaspectratio]{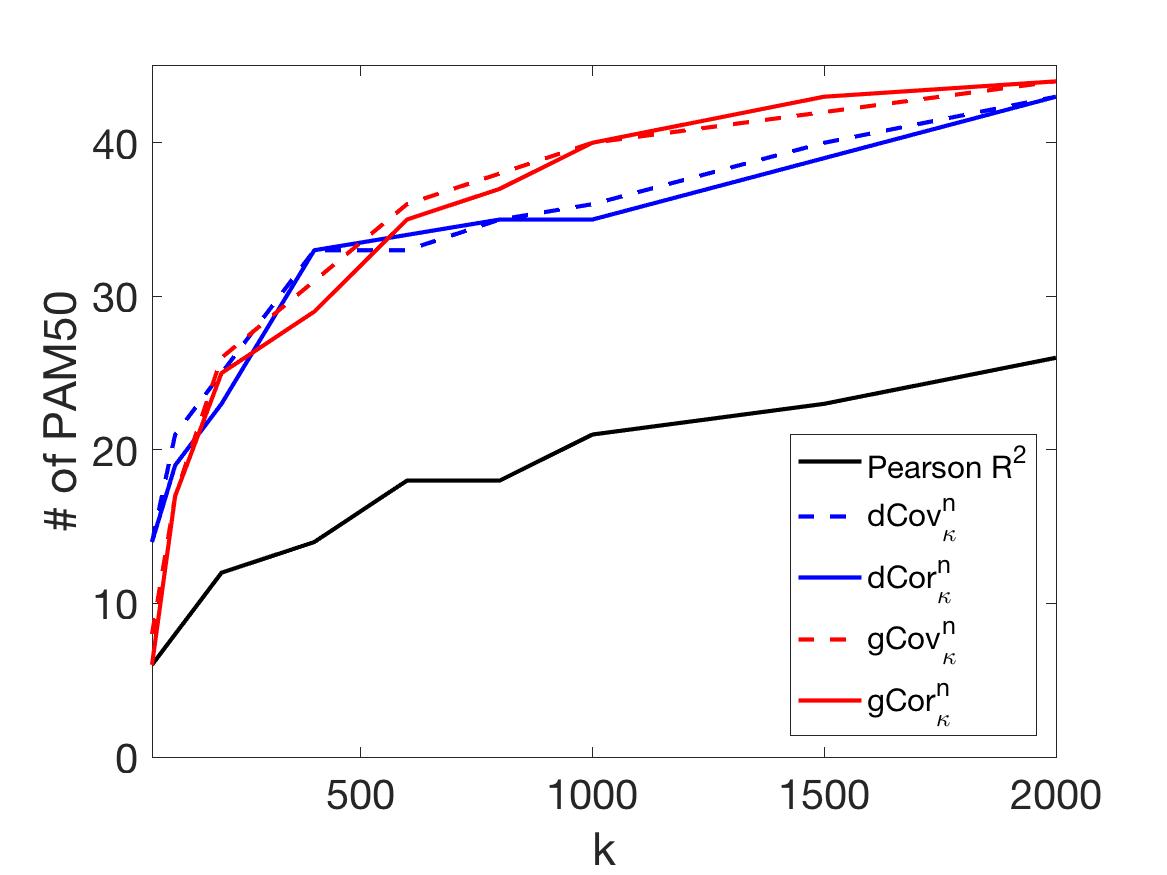} \\
    (a) & \hskip -.4in (b) 
\end{tabular}
\caption{The breast cancer dataset. (a) Test accuracy using the top $k$ selected genes. (b) Number of PAM50 genes in the selected top $k$ genes.} 
\label{figure:BRCA} 
\end{figure*}

The results are shown in Fig.~\ref{figure:BRCA}. Fig.~\ref{figure:BRCA}(a) shows the classification performance using selected top $k$ genes using different test statistics as well as using all PAM50 genes (shown as a green dotted line). Note that the accuracy of PAM50 is one averaged value from 30 runs. We plot it as a line across the entire $k$ range for easier comparison with other methods. Among the 5 selection methods under comparison, $\mathrm{gCov}_\kappa^n$ has the best overall performance, even outperforms PAM50 with $k=30$. This suggests that $\mathrm{gCov}_\kappa^n$ is able to select a smaller number of genes and the prediction is better than the gold standard. We also observe that $\mathrm{gCor}_\kappa^n$ outperforms PAM50 with $k=40$ and $k=50$. Pearson $R^2$, $\mathrm{dCov}_\kappa^n$, and $\mathrm{gCor}_\kappa^n$ are not able to exceed PAM50 within 50 genes. Fig.~\ref{figure:BRCA}(b) shows the number of PAM50 genes appear in the top $k$ selected genes for each selection method. It is obvious to see that Pearson $R^2$ selects the smallest number of PAM50. Among the top 2000 genes selected by Pearson $R^2$, only half of the PAM50 gene are included. Gini statistics are able to select more PAM50 genes than distance statistics as $k$ increases. The small ratio of PAM50 included in the selected genes by any method is because of the high correlation between genes. PAM50 was derived by not only selecting most subtype dependent genes, but also less mutually dependent genes to obtain a smaller set of genes for the same prediction accuracy. Even though any of the selection methods under comparison does not take the feature-feature dependence into consideration, both $\mathrm{gCov}_\kappa^n$ and $\mathrm{gCor}_\kappa^n$ are able to select a better gene set than PAM50 for classification.

\subsection{UCI Datasets}
We further tested the 5 feature selection methods on a total of 15 UCI datasets of classification tasks using numerical features. The 15 datasets cover a wide range on both the sample and feature set size. Specifically, we avoided binary-class and balanced datasets because $\mathrm{dCov}_\kappa^n$ and $\mathrm{gCov}_\kappa^n$ give the same ranking on these datasets when sufficient training samples are given. For datasets without training and test sets provided, we randomly hold out 20\% as the test set. The descriptions of these datasets used are summarized in TABLE~\ref{table_UCI_summary}. For the MNIST and UJInddorLoc datasets, we randomly selected 5000 samples from the training set to calculate test statistics. For the remaining datsets, all training samples were used. For each dataset, each method was used to select top $k$ features for training with three different values of $k$. Each classification test was repeated 10 times. Other experimental setups were kept the same as previously described. 

As we do not have the ground truth of the dependent features for most of these datasets, only classification accuracy was used for evaluation. The average test accuracies (from 10 runs) of the 5 methods under comparison with different values of $k$ on the 15 UCI datsets are summarized in TABLE~\ref{table_UCI_results}. Of the 5 methods, the highest accuracy is shown in bold and the second highest one is underlined. The top 1 statistic is calculated by how many times a method is shown in bold and the top 2 statistic is obtained by the total number of times a method is shown in bold or underlined. Among all methods, $\mathrm{gCor}_\kappa^n$ appears 18 times as top 1 and 28 times in top 2, outperforming all other methods. The second best method is $\mathrm{dCov}_\kappa^n$, followed by $\mathrm{gCov}_\kappa^n$ and then $\mathrm{dCor}_\kappa^n$. Peasron $R^2$ is still the worst, with only 8 times as top 1 and 14 time in top 2. Comparing TABLE~\ref{table_UCI_results} and Fig.~\ref{figure:mnist}(b), we see no correlation between sensitivity and ranking performance of test statistics. The ranking performance is data dependent and $\mathrm{gCor}_\kappa^n$ performances better on average.

\begin{table}[t]
\caption{Data Sets Summary} \label{table_UCI_summary}
\centering
\begin{tabular}{ ||lcccc||}
\hline\hline
Data Set & Train Size & Test Size & Features & Classes \\
\hline\hline
MNIST & 60000 & 10000 & 784 & 10 \\
Breast Cancer & 413 & 103 & 17278 & 5 \\
Gene Expression & 641 & 160 & 20531 & 5 \\
Gastro Lesions & 61 & 15 & 1396 & 3 \\
Satellite & 4435 & 2000 & 36 & 6 \\
Ecoli & 269 & 67 & 7 & 8 \\
Glass & 171 & 43 & 9 & 6 \\
Urban Land & 168 & 507 & 147 & 9 \\
Wine & 142 & 36 & 13 & 3 \\
Anuran Calls & 5756 & 1439 & 22 & 4 \\
Breast Tissue & 85 & 21 & 9 & 4 \\
Cardiotocography & 1701 & 425 & 21 & 10 \\
Leaf & 272 & 68 & 14 & 30 \\
Mice Protein Exp & 864 & 216 & 77 & 8 \\
HAR & 4252 & 1492 & 561 & 6 \\
UJIndoorLoc & 19937 & 1111 & 520 & 13 \\
Forest Types & 198 & 325 & 27 & 4 \\
\hline\hline
\end{tabular}
\end{table}

\begin{table*}
\caption{Classification Accuracies Using Top k Dependent Features.} \label{table_UCI_results}
\centering
\begin{tabular}{ ||l||ccc|ccc|ccc|ccc|| }
\hline\hline
&\multicolumn{3}{c|}{Gene Expression}&\multicolumn{3}{c|}{Gastrointestinal Lesions}&\multicolumn{3}{c|}{Satellite}&\multicolumn{3}{c||}{Ecoli} \\
\hline
k & 5 & 7 & 10 & 10 & 100 & 200 & 5 & 10 & 15 & 2 & 3 & 4 \\
\hline
Pearson $R^2$ & \underline{0.909} & 0.917 & 0.934 & \underline{0.747} & \textbf{0.700} & 0.660 & 0.598 & 0.830 & 0.869 & 0.624 & \textbf{0.772} & 0.776 \\
$\mathrm{dCov}_\kappa^n$ & 0.889 & 0.907 & \underline{0.963} & 0.700 & 0.607 & 0.673 & 0.627 & 0.860 & 0.896 & 0.712 & \underline{0.766} & \underline{0.788} \\
$\mathrm{dCor}_\kappa^n$ & 0.851 & 0.928 & 0.937 & \textbf{0.787} & 0.633 & \underline{0.680} & 0.808 & \textbf{0.887} & \textbf{0.903} & \underline{0.713} & 0.764 & 0.787 \\
\hline
$\mathrm{gCov}_\kappa^n$ & \textbf{0.923} & \underline{0.933} & 0.948 & 0.560 & \underline{0.647} & \textbf{0.693} & \underline{0.834} & 0.870 & 0.881 & \textbf{0.718} & \underline{0.766} & 0.781 \\
$\mathrm{gCor}_\kappa^n$ & 0.741 & \textbf{0.980} & \textbf{0.972} & 0.607 & \textbf{0.700} & 0.653 & \textbf{0.838} & \underline{0.873} & \underline{0.898} & 0.634 & 0.758 & \textbf{0.803} \\
\hline\hline
&\multicolumn{3}{c|}{Glass}&\multicolumn{3}{c|}{Urban Land Cover}&\multicolumn{3}{c|}{Wine}&\multicolumn{3}{c||}{Anuran Calls} \\
\hline
k & 3 & 5 & 7 & 30 & 60 & 90 & 2 & 4 & 6 & 5 & 10 & 15 \\
\hline
Pearson $R^2$ & 0.702 & \textbf{0.730} & 0.707 & 0.764 & 0.778 & 0.798 & 0.753 & 0.969 & 0.972 & 0.927 & \underline{0.957} & 0.975 \\
$\mathrm{dCov}_\kappa^n$ & 0.702 & \underline{0.691} & \underline{0.744} & \underline{0.786} & 0.795 & \textbf{0.809} & \underline{0.897} & \textbf{0.997} & \textbf{1.000} & \textbf{0.938} & \underline{0.957} & \underline{0.979} \\
$\mathrm{dCor}_\kappa^n$ & \textbf{0.707} & 0.672 & \textbf{0.751} & 0.781 & \textbf{0.817} & \textbf{0.809} & 0.881 & \underline{0.992} & \underline{0.997} & \underline{0.937} & 0.956 & 0.978 \\
\hline
$\mathrm{gCov}_\kappa^n$ & \underline{0.705} & 0.674 & 0.663 & 0.784 & 0.790 & 0.804 & 0.881 & \textbf{0.997} & \underline{0.997} & \textbf{0.938} & 0.956 & \underline{0.979} \\
$\mathrm{gCor}_\kappa^n$ & 0.693 & 0.665 & 0.670 & \textbf{0.787} & \underline{0.804} & \underline{0.805} & \textbf{0.900} & \textbf{0.997} & \textbf{1.000} & \underline{0.937} & \textbf{0.958} & \textbf{0.980} \\
\hline\hline
&\multicolumn{3}{c|}{Breast Tissue}&\multicolumn{3}{c|}{Cardiotocography}&\multicolumn{3}{c|}{Leaf}&\multicolumn{3}{c||}{Mice Protein Expression} \\
\hline
k & 3 & 5 & 7 & 5 & 10 & 15 & 4 & 7 & 10 & 10 & 20 & 30 \\
\hline
Pearson $R^2$ & \underline{0.810} & \textbf{0.857} & 0.833 & \textbf{0.831} & \textbf{0.890} & \underline{0.894} & 0.437 & 0.637 & 0.647 & \textbf{0.978} & 0.942 & 0.980 \\
$\mathrm{dCov}_\kappa^n$ & \textbf{0.819} & \textbf{0.857} & \underline{0.852} & 0.773 & 0.874 & 0.891 & 0.471 & \textbf{0.690} & 0.663 & \underline{0.893} & \underline{0.950} & \textbf{0.985} \\
$\mathrm{dCor}_\kappa^n$ & \underline{0.810} & \textbf{0.857} & 0.838 & 0.816 & 0.849 & \underline{0.894} & \textbf{0.506} & 0.606 & \underline{0.704} & 0.890 & \textbf{0.952} & 0.977 \\
\hline
$\mathrm{gCov}_\kappa^n$ & \underline{0.810} & \textbf{0.857} & \underline{0.852} & \underline{0.817} & 0.878 & \textbf{0.895} & 0.468 & \underline{0.688} & 0.654 & 0.888 & 0.945 & \underline{0.983} \\
$\mathrm{gCor}_\kappa^n$ & \underline{0.810} & \textbf{0.857} & \textbf{0.857} & 0.766 & \underline{0.880} & 0.887 & \underline{0.503} & 0.594 & \textbf{0.706} & 0.888 & 0.943 & 0.982 \\
\hline\hline
&\multicolumn{3}{c|}{HAR}&\multicolumn{3}{c|}{UJIndoorLoc}&\multicolumn{3}{c|}{Forest Types}& Top 1 & Top 2 & \\
\cline{1-10}
k & 100 & 200 & 300 & 100 & 200 & 300 & 5 & 10 & 15 & (times) & (times) & \\
\hline
Pearson $R^2$ & 0.756 & \underline{0.780} & 0.856 & 0.695 & 0.811 & 0.847 & \textbf{0.750} & 0.758 & 0.795 & 8 & 14 & \\
$\mathrm{dCov}_\kappa^n$ & 0.765 & 0.779 & \textbf{0.859} & 0.768 & \textbf{0.866} & \textbf{0.873} & 0.653 & \underline{0.810} & \underline{0.804} & 11 & 25 & \\
$\mathrm{dCor}_\kappa^n$ & \underline{0.767} & \underline{0.780} & 0.853 & \underline{0.793} & 0.863 & 0.871 & \underline{0.715} & 0.756 & \underline{0.804} & 10 & 23 & \\
\hline
$\mathrm{gCov}_\kappa^n$ & 0.766 & \textbf{0.853} & \underline{0.858} & 0.763 & \underline{0.864} & \underline{0.872} & 0.651 & \textbf{0.817} & \textbf{0.805} & 10 & 25 & \\
$\mathrm{gCor}_\kappa^n$ & \textbf{0.821} & \textbf{0.853} & 0.853 & \textbf{0.797} & \underline{0.864} & \underline{0.872} & 0.695 & 0.735 & \textbf{0.805} & \textbf{18} & \textbf{28} & \\
\hline\hline
\end{tabular}
\end{table*}

\section{Conclusions}\label{conclusions}

We have proposed a feature selection framework based on a new dependence measure between a numerical feature $X$ and a categorical label $Y$ using generalized Gini distance statistics: Gini distance covariacne $\mathrm{gCov}(X,Y)$ and Gini distance correlation $\mathrm{gCor}(X,Y)$. We presented estimators of $\mathrm{gCov}(X,Y)$ and $\mathrm{gCor}(X,Y)$ using $n$ iid samples, i.e., $\mathrm{gCov}_\kappa^n$ and $\mathrm{gCor}_\kappa^n$, and
derived uniform convergence bounds. We showed that $\mathrm{gCov}_\kappa^n$ converge faster than its distance statistic counterpart $\mathrm{dCov}_\kappa^n$, and the probability of $\mathrm{gCov}_\kappa^n$ under-performing $\mathrm{dCov}_\kappa^n$ in Type II error decreases to 0 exponentially as the sample size increases. $\mathrm{gCov}_\kappa^n$ and $\mathrm{gCor}_\kappa^n$ are also simpler to calculate than $\mathrm{dCov}_\kappa^n$ and $\mathrm{dCor}_\kappa^n$. Extensive experiments were performed to compare $\mathrm{gCov}_\kappa^n$ and $\mathrm{gCor}_\kappa^n$ with other dependence measures in feature selection tasks using artifical and real datasets, including MNIST, breast cancer and 15 UCI datsets. For simulated datasets, $\mathrm{gCov}_\kappa^n$ and $\mathrm{gCor}_\kappa^n$ perform better in terms of power and AUC. For real datasets, on average, $\mathrm{gCov}_\kappa^n$ and $\mathrm{gCor}_\kappa^n$ are able to select more meaningful features and have better classification performances.

The proposed feature selection method using generalized Gini distance statistics have several limitations:
\begin{itemize}
    \item Choosing an optimal $\sigma$ is still an open problem. In our experiments we used $\sigma^2=10$ after data standardization;
    \item The computation cost for $\mathrm{gCov}_\kappa^n$ and $\mathrm{gCor}_\kappa^n$ is $O(n^2)$, which is same as $\mathrm{dCov}_\kappa^n$ and $\mathrm{dCor}_\kappa^n$, but more expensive than Pearson $R^2$ which takes linear time. For large datasets, a sampling of data is desired.
    \item Features selected by Gini distance statistics, as well as other dependence measure based methods, can be redundant,  hence a subsequent feature elimination may be needed for the sake of feature subset selection.
\end{itemize}

%

\newpage
\ifCLASSOPTIONcompsoc
  
  \section*{Acknowledgments}
\else
  \section*{Acknowledgment}
\fi

This work was supported by the Big Data Constellation of the University of Mississippi.

\ifCLASSOPTIONcaptionsoff
  \newpage
\fi



%

%

\begin{IEEEbiography}{Silu Zhang}
received the BS degree in bioengineering from Zhejiang University, China and the MS degree in chemical engineering from North Carolina State University, both in 2011. She is currently a PhD candidate in computer science at the University of Mississippi, working on machine learning research in feature selection and improving random forest, as well as application of machine learning and statistical models on gene expression data.
\end{IEEEbiography}

\begin{IEEEbiography}{Xin Dang}
received the BS degree in Applied Mathematics (1991) from Chongqing University, China, the Master degree (2003) and the PhD degree (2005) in Statistics from the University of Texas at Dallas. Currently she is a professor of
the Department of Mathematics at the University
of Mississippi. Her research interests include robust
and nonparametric statistics, statistical and numerical computing, and multivariate data analysis. In
particular, she has focused on data depth and applications, bioinformatics, machine learning, and robust
procedure computation. Dr. Dang is a member of the
IMS, ASA, ICSA and IEEE.

\end{IEEEbiography}

\begin{IEEEbiography}{Dao Nguyen} received Bachelor of Computer Science (1997) from the University of Wollongong, Australia, Ph.D. of Electrical Engineering (2010) from the University of Science and Technology, Korea. In 2016, he received the Ph.D. degree in Statistics from The University of Michigan, Ann Arbor. He was a postdoc scholar at the University of California, Berkeley for more than a year before becoming an Assistant Professor of Mathematics at the University of Mississippi in 2017. His research interests  include machine learning, dynamics modeling, stochastic optimization, Bayesian analysis. Dr. Nguyen is a member of the ASA, ISBA. 
\end{IEEEbiography}

\begin{IEEEbiography}{Dawn Wilkins}
received the B.A. and M.A. degrees in Mathematical Systems from Sangamon State University (now University of Illinois--Springfield) in 1981 and 1983, respectively.  After earning a Ph.D. in Computer Science from Vanderbilt University in 1995, she joined the faculty of the University of Mississippi, where she is 
now Professor of Computer and Information Science.  Dr. Wilkins' research interests are primarily in machine learning,
computational biology, bioinformatics and database systems.
\end{IEEEbiography}

\begin{IEEEbiography}{Yixin Chen} received B.S.degree (1995) from the Department of Automation, Beijing Polytechnic University, the M.S. degree (1998) in control theory and application from Tsinghua University, and the M.S. (1999) and Ph.D. (2001) degrees in electrical engineering from the University  of Wyoming. In 2003, he received the Ph.D. degree in computer science from The Pennsylvania State University. He had been an Assistant Professor of computer science at University of New Orleans. He is now a Professor of Computer and Information Science at the University of Mississippi. His research interests  include machine learning, data mining, computer  vision, bioinformatics, and robotics and control. Dr. Chen is a member of the ACM, the IEEE, the IEEE Computer Society, and the IEEE Neural Networks Society. 
\end{IEEEbiography}








\end{document}


\title{Estimating Feature-Label Dependence Using Gini Distance Statistics}

\author{Silu Zhang,
        Xin Dang,~\IEEEmembership{Member,~IEEE,}
        Dao Nguyen,
        Dawn Wilkins,~Yixin Chen,~\IEEEmembership{Member,~IEEE}
}

\markboth{IEEE Transactions on Pattern Analysis and Machine Intelligence,~Vol.~, No.~, ~2019}
{Estimating Feature-Label Dependence Using Gini Distance Statistics}
%

\maketitle


%

\section*{SUPPLEMENTARY MATERIAL}
\begin{figure*}
\hskip -.18in
\centering
\begin{tabular}{ccc}
    \includegraphics[draft=false, width = 2.5in, keepaspectratio]{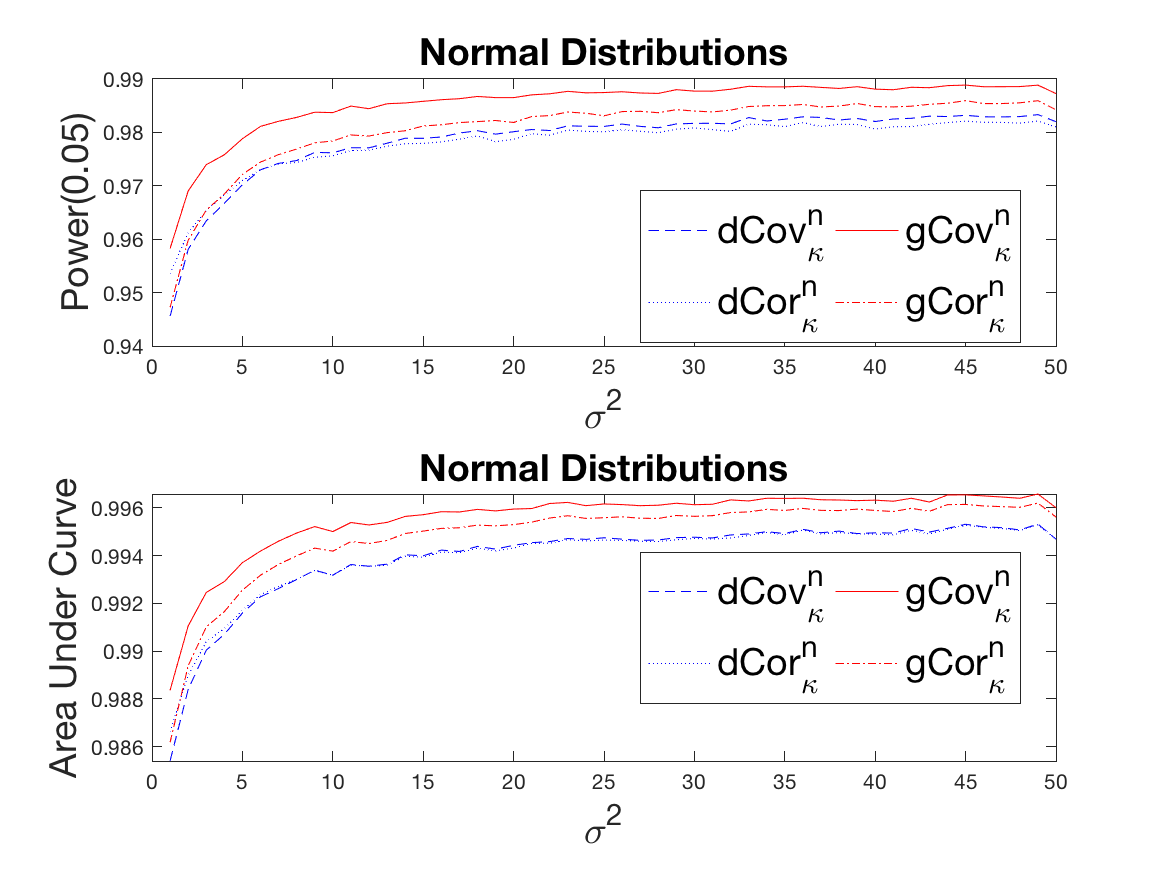}
    & \hskip -.37in \includegraphics[draft=false, width = 2.5in, keepaspectratio]{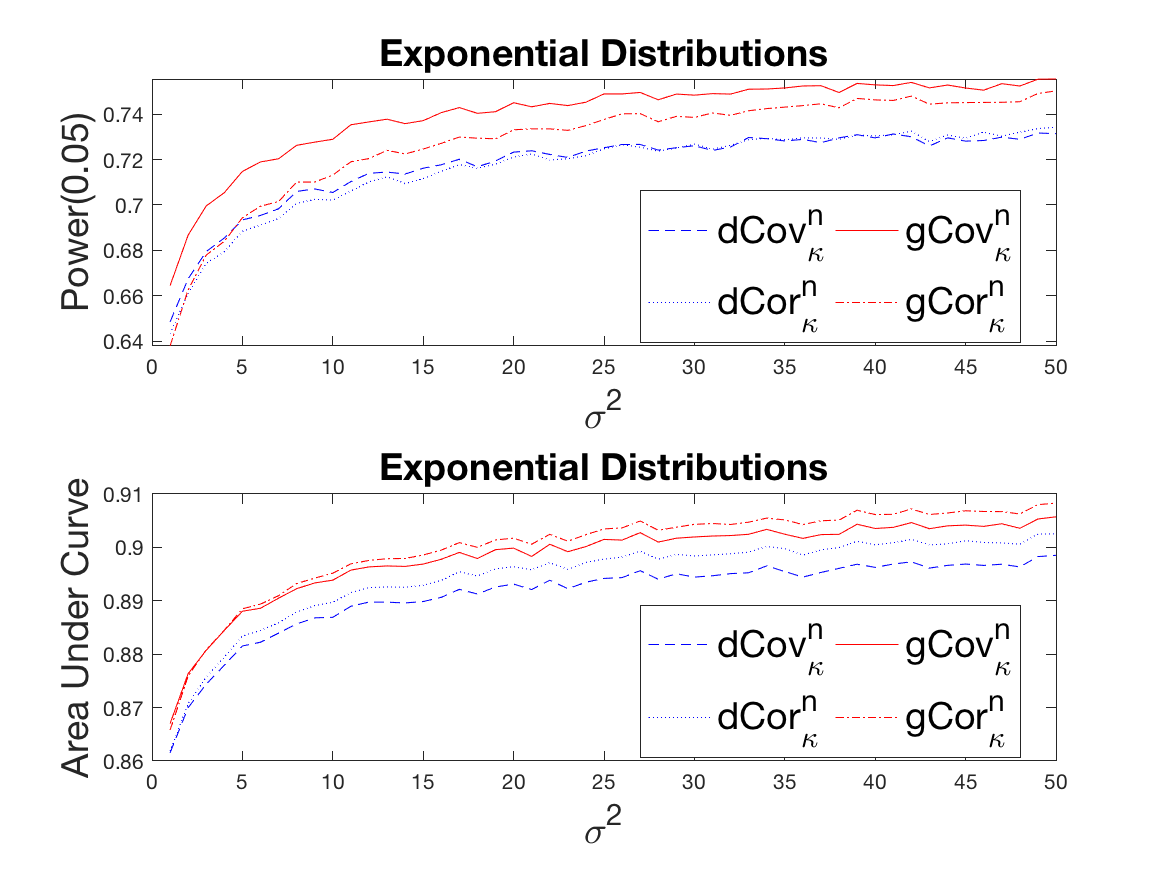} & \hskip -.37in \includegraphics[draft=false, width = 2.5in, keepaspectratio]{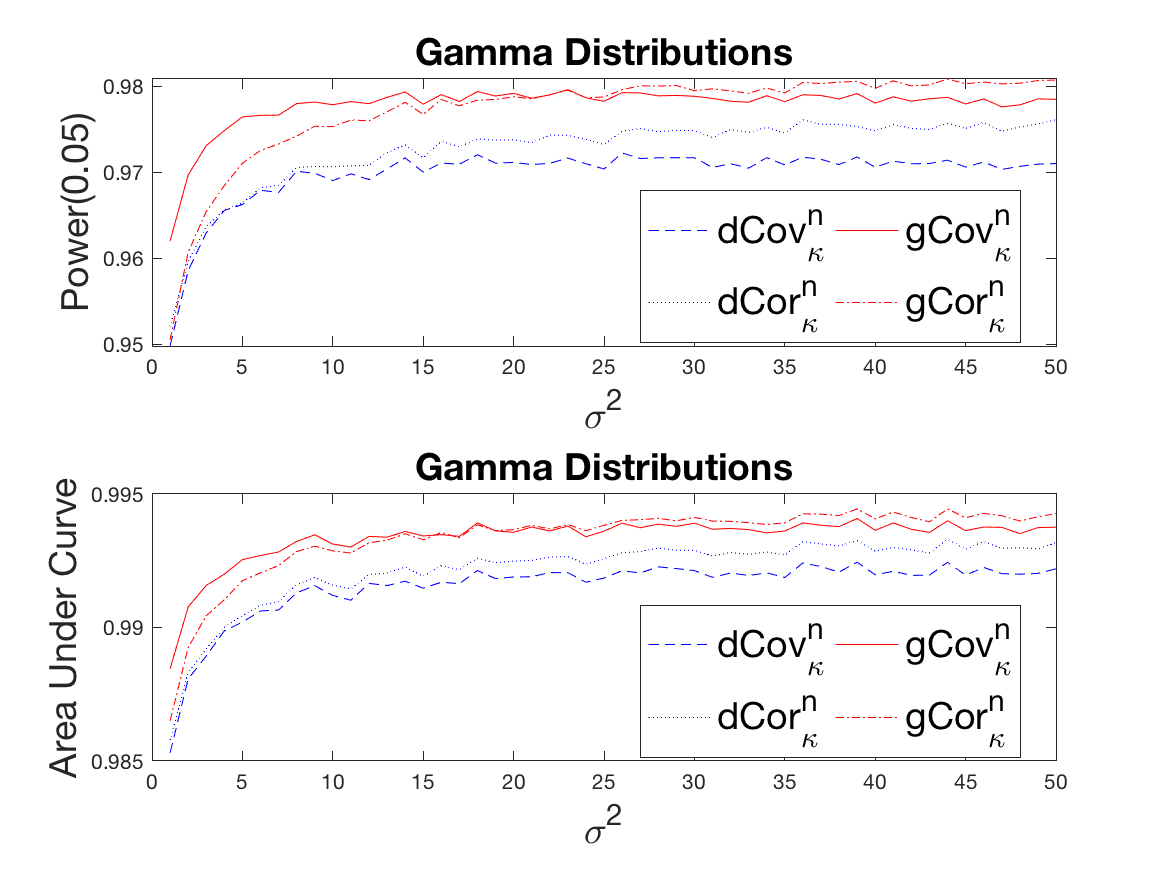}\\
     & \hskip -.37in (a) $K=3$ & \hskip -.37in \\
    \includegraphics[draft=false, width = 2.5in, keepaspectratio]{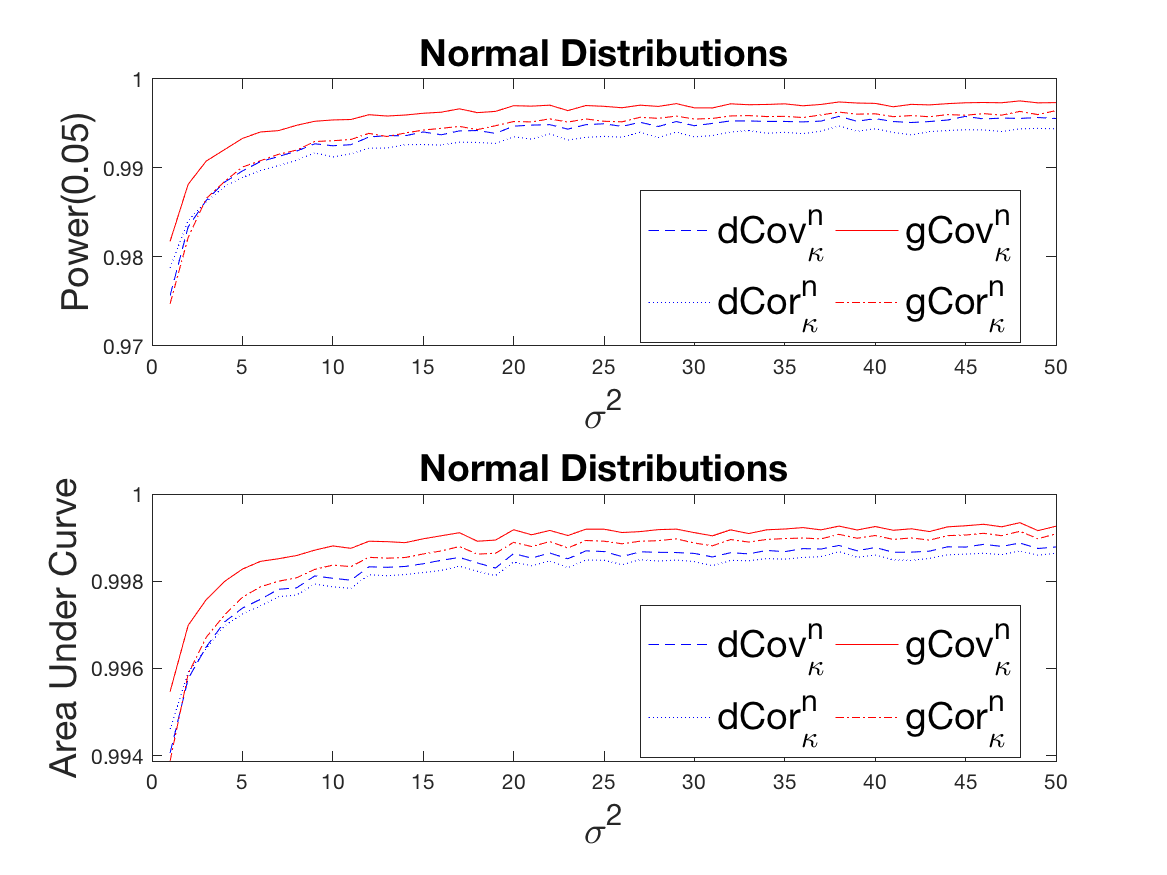}
    & \hskip -.37in \includegraphics[draft=false, width = 2.5in, keepaspectratio]{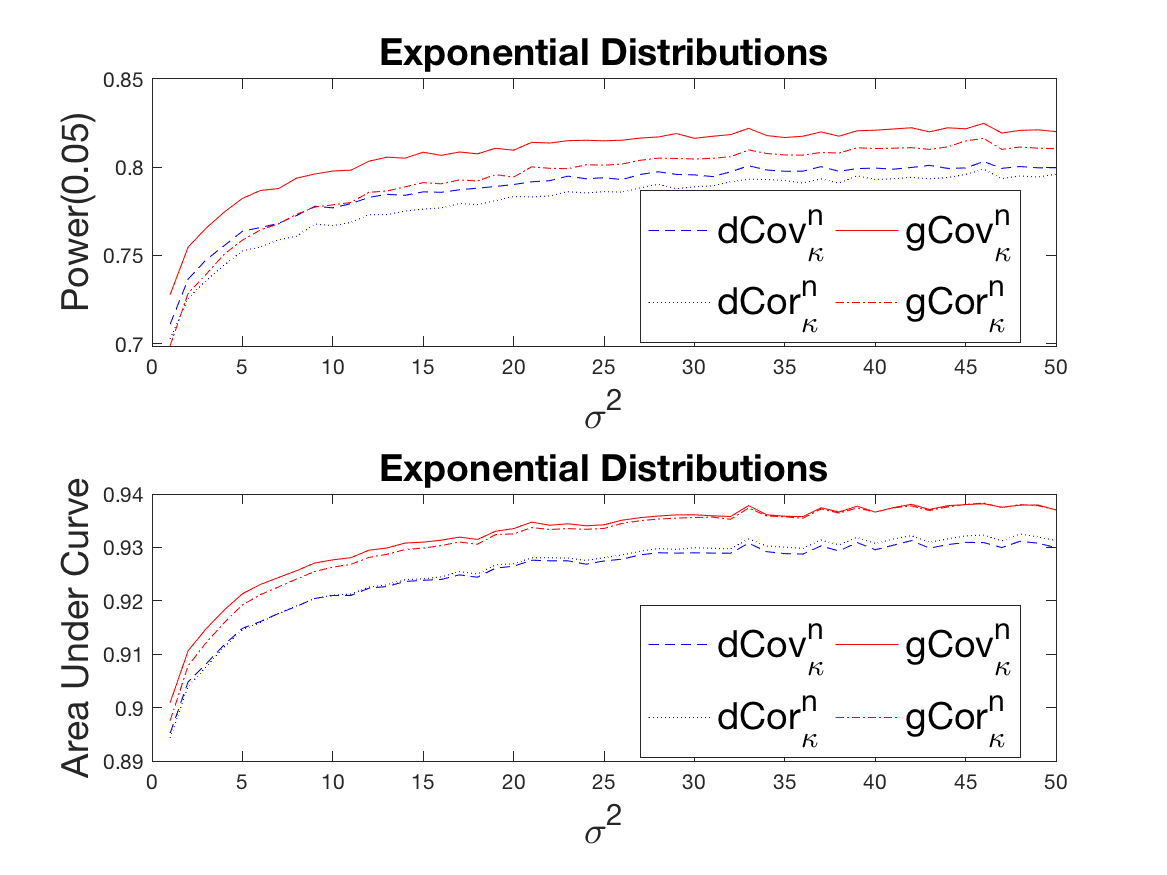} & \hskip -.37in \includegraphics[draft=false, width = 2.5in, keepaspectratio]{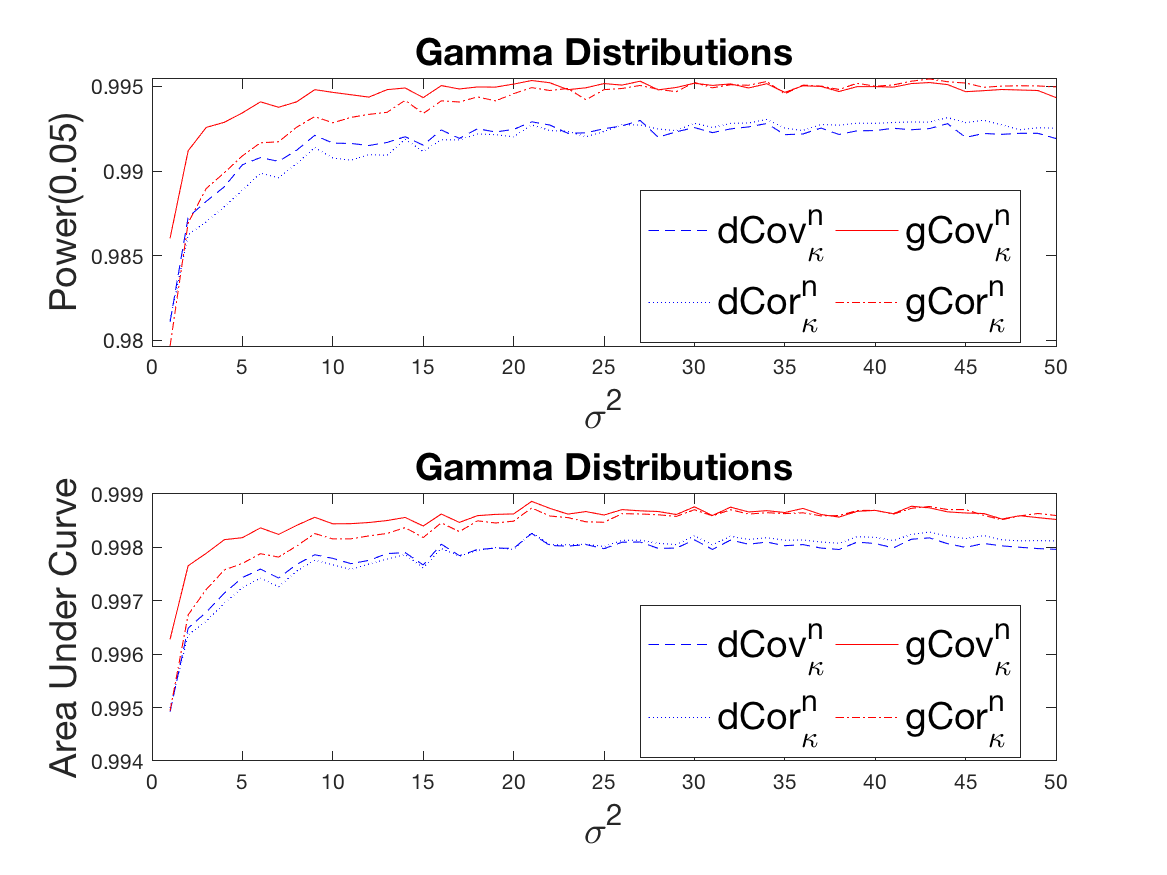}\\
     & \hskip -.37in (b) $K=4$ & \hskip -.37in \\
    \includegraphics[draft=false, width = 2.5in, keepaspectratio]{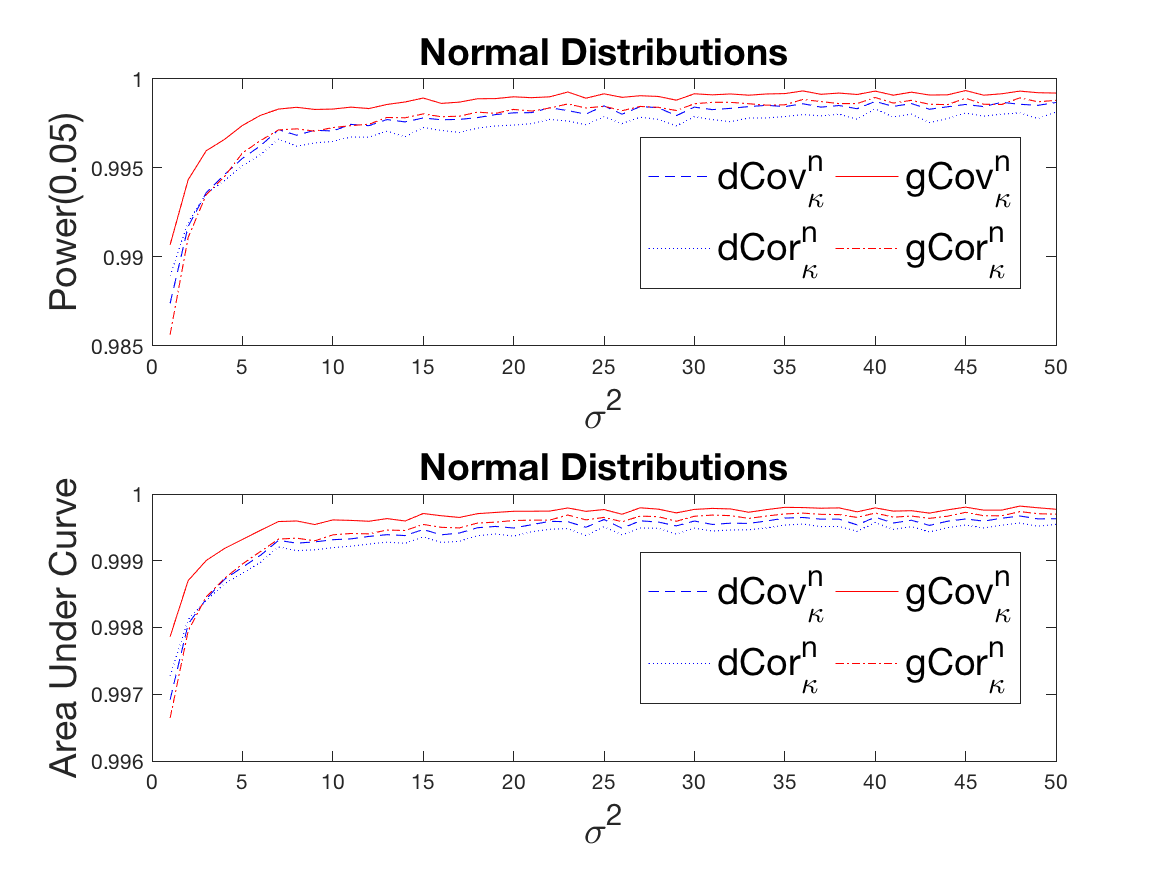}
    & \hskip -.37in \includegraphics[draft=false, width = 2.5in, keepaspectratio]{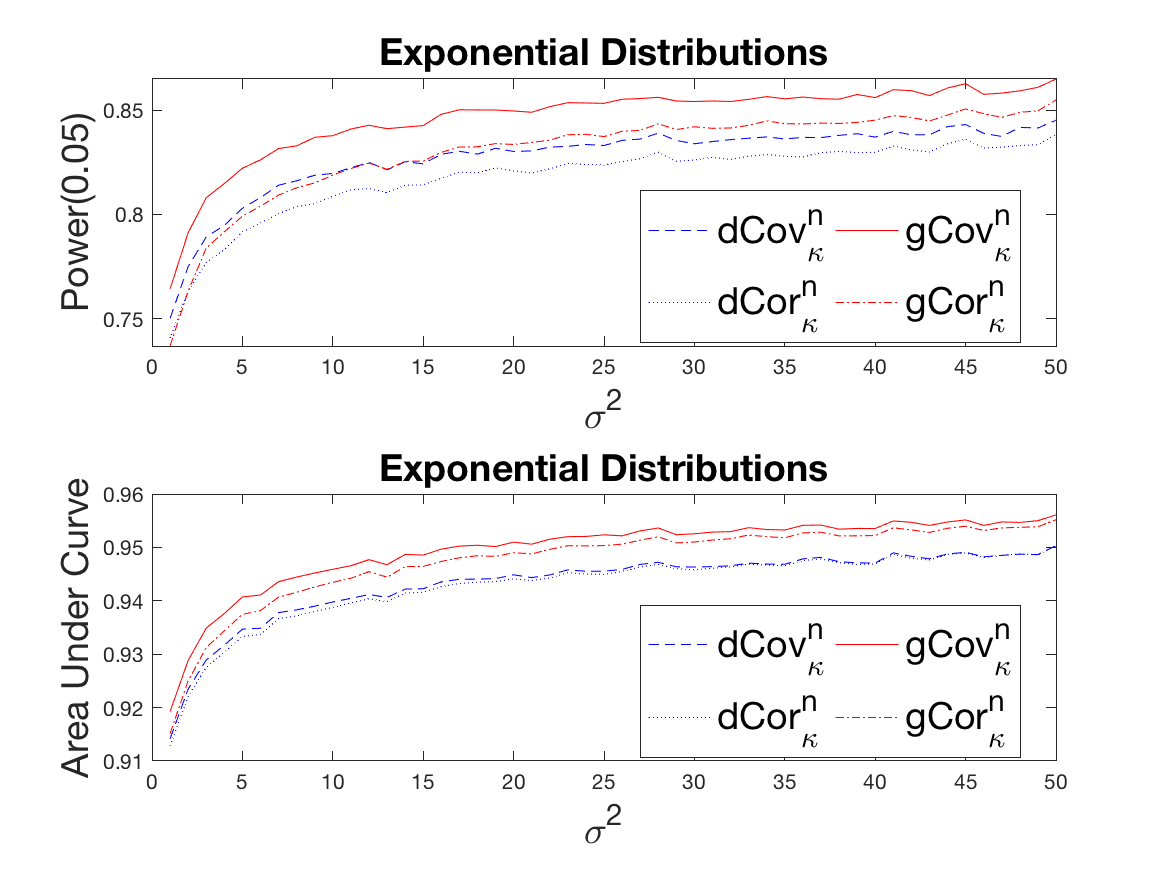} & \hskip -.37in \includegraphics[draft=false, width = 2.5in, keepaspectratio]{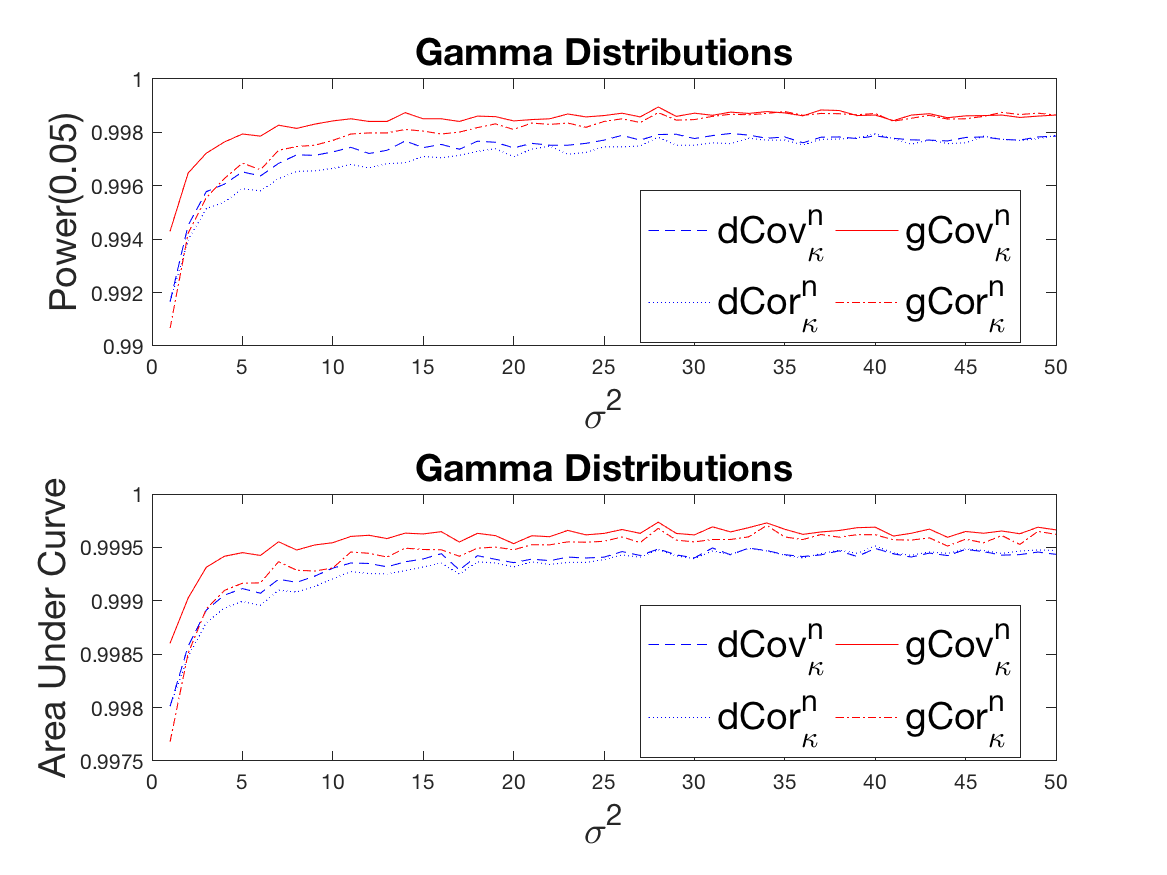}\\
     & \hskip -.37in (c) $K=5$ & \hskip -.37in
\end{tabular}
\caption{Power of test statistics at $0.05$ Type I error and AUC. Left: Normal distribution family, Middle: Exponential distribution family, Right: Gamma distribution family.} \label{figure_power_auc}
\end{figure*}

This document presents the effect of the kernel parameter $\sigma$ on performance of the four test statistics under different values of K for three distribution families, as shown in Figure~\ref{figure_power_auc}. The performance of each method becomes more insensitive to $\sigma$ as $\sigma$ increases and a larger $\sigma$ is more favorable. The relative performances of the four methods are barely affected by the choice of $\sigma$.
